\documentclass[journal]{IEEEtran}

\usepackage[linesnumbered,ruled,vlined]{algorithm2e}
\usepackage{multirow,booktabs,color,soul,threeparttable,longtable}
\usepackage{booktabs}
\usepackage{multirow}
\usepackage{subfigure}
\usepackage{epsfig}
\usepackage{amsmath}
\usepackage{amssymb}
\usepackage{lineno,hyperref}
\usepackage{setspace}
\usepackage{amsfonts}
\usepackage{url}

\ifCLASSINFOpdf
\else
\fi

\begin{document}

\title{Integrating Variable Reduction Strategy with Evolutionary Algorithm for Solving Nonlinear Equations Systems}

\author{Aijuan Song, Guohua Wu, and Witold Pedrycz,~\IEEEmembership{Fellow,~IEEE}

\thanks{Aijuan Song and Guohua Wu are with the School of Traffic and Transportation Engineering, Central South University, Changsha 410075, China. E-mail: aijuansong@csu.edu.cn; guohuawu@csu.edu.cn.}

\thanks{Witold Pedrycz is with the Department of Electrical and Computer Engineering, University of Alberta, Edmonton, AB T6G 2V4, Canada, with the Department of Electrical and Computer Engineering, Faculty of Engineering, King Abdulaziz University, Jeddah 21589, Saudi Arabia, and also with the Systems Research Institute, Polish Academy of Sciences, Warsaw 01447, Poland. E-mail: wpedrycz@ualberta.ca.}

}


\maketitle

\begin{abstract}

Nonlinear equations systems (NESs) are widely used in real-world problems while they are also difficult to solve due to their characteristics of nonlinearity and multiple roots. Evolutionary algorithm (EA) is one of the methods for solving NESs, given their global search capability and an ability to locate multiple roots of a NES simultaneously within one run. Currently, the majority of research on using EAs to solve NESs focuses on transformation techniques and improving the performance of the used EAs. By contrast, the problem domain knowledge of NESs is particularly investigated in this study, using which we propose to incorporate the variable reduction strategy (VRS) into EAs to solve NESs. VRS makes full use of the systems of expressing a NES and uses some variables (i.e., core variable) to represent other variables (i.e., reduced variables) through the variable relationships existing in the equation systems. It enables to reduce partial variables and equations and shrink the decision space, thereby reducing the complexity of the problem and improving the search efficiency of the EAs. To test the effectiveness of VRS in dealing with NESs, this paper integrates VRS into two existing state-of-the-art EA methods (i.e., MONES and DR-JADE), respectively. Experimental results show that, with the assistance of VRS, the EA methods can significantly produce better results than the original methods and other compared methods.

\end{abstract}


\IEEEpeerreviewmaketitle

\section{Introduction}

\IEEEPARstart{N}{onlinear} equations systems (NESs) have emerged in fields of science, engineering, economics, etc \cite{mehta2015collection}. Numerous real-world problems can be modeled as NESs such as chemistry \cite{holstad1999numerical}, robotics \cite{collins2002forward}, electronic \cite{facchinei2007generalized}, signal processing \cite{chaudhary2017modified}, and physics \cite{yuan2008new}. Unlike linear equations systems, NESs have plenty of nonlinear operators like ${{\rm{a}}^x}$, $\ln x$, ${x^{\rm{n}}}$ and trigonometric function. Furthermore, the overwhelming majority of NESs have more than a single equally important root. The above two characteristics make the problem challenging.

There is a class of methods for finding the numerical solutions of NESs. This kind of method mainly includes Newton methods \cite{karr1998solutions}, quasi-Newton methods \cite{karr1998solutions}, interval-based methods, e.g., interval-Newton \cite{grosan2008new}, homotopy continuation (embedding) methods \cite{grosan2008new, bates2013numerically}, trust-region method \cite{bates2013numerically}, secant method  \cite{denis1993least}, Halley method \cite{ortega1970iterative}, branch and bound method \cite{ortega1970iterative}, etc.. Nevertheless, a host of these methods have the following weaknesses: a) Aiming at locating just a single optimal solution rather than multiple optimal solutions in a single run; b) High requirements for prior knowledge, such as derivative, good starting value; c) The quality of solutions is strongly problem-dependent and depends on the initial guess. Therefore, such methods are no longer completely sufficient. Over the past decade, people have also developed metaheuristics to solve NESs while continuously improving these methods to study the numerical solution of NESs.

The metaheuristics possess advantages such as simple implementation, strong versatility and strong global search capability \cite{wang2001intelligent}. In comparison with the methods to study the numerical solution of NESs, it enjoys unique benefits especially for solving complex problems such as less prior knowledge required, no derivative requirement, and less dependence on problem characteristics and the initial solution. However, metaheuristics also face quite a few challenges when solving NESs, for instance, when dealing with large-scale and high-dimensional NESs and finding multiple roots in a single run.

Evolutionary algorithm (EA) is essentially a global stochastic search algorithm \cite{wang2001intelligent} and demonstrates the capacity for locating multiple solutions over a single run. EAs are prevalent and effective methods for solving NESs. A NES needs to be transformed into an optimization problem prior to the solving process by EAs. The transformed problem is generally a multi-modal or multi-objective optimization problem. At present, EAs that have been applied to solve NESs comprise evolutionary strategy (ES), particle swarm optimization algorithm (PSO), differential evolution algorithm (DE), genetic algorithm (GA), etc.. For example, Tong et al. proposed a ranking method in ES for solving NESs \cite{geng2009research}. Ouyang et al. developed a hybrid PSO \cite{ouyang2009hybrid}, which solves NESs by combining PSO with the Nelder-Mead simplex method. Turgut et al. designed a chaotic behavior PSO to solve NESs \cite{turgut2014chaotic}, which improves the robustness and effectiveness of the algorithm through different chaotic maps. Gong et al. argued that locating multiple roots by repulsion techniques is a promising method and they proposed a repulsion-based adaptive DE (RADE) for solving NESs \cite{gong2018finding}. Ren et al. developed an efficient GA with symmetric and harmonious individuals for solving NESs \cite{ren2013solving} and Joshi et al. used an improved GA to solve NESs \cite{joshi2014solving}. Thereby, obtaining multiple optimization solutions for the optimization problem by EAs corresponds to getting multiple roots of the NES.

To solve NESs by EAs effectively, we can pay attention to the transformation technique, the algorithm and the problem itself, i.e., designing more reliable transformation techniques, designing more efficient EAs and reducing the complexity of the problem. Some research shows that effectively integrating an algorithm with problem domain knowledge can generally improve the performance of the algorithm \cite{beyer2002evolution}. Unfortunately, at present, for solving NES the main research dedicates to transformation techniques and the performance of EAs while lacking relevant research on the complexity reduction of problems.

VRS can make full use of the problem domain knowledge to reduce problems and trigger the complexity reduction of problems. Currently, VRS has been applied to equality constrained optimization problems \cite{wu2015variable} and derivative unconstrained optimization problems \cite{wu2017using}, which have significantly improved the optimization efficiency. Therefore, it is of considerable significance to study how to apply VRS to solve NESs effectively.
Based on the above considerations, we propose to integrate VRS with EAs for solving NESs. VRS represents some variables with other variables through the relationships among variables in the equations, resulting in reducing the complexity of the problems and improving the search efficiency of the algorithm. The main contributions of this paper can be summarized as follows:

\begin{itemize}
	
	\item We propose to utilize VRS to reduce the complexity of NESs. We elaborately analyze and explain how to apply VRS to simplify a NES. With the assistance of VRS, a NES can entail the smaller decision space and lower complexity.

	\item A general framework is proposed for integrating VRS with an arbitrary EA for solving NESs. By this framework, the optimization efficiency of the used EAs can be significantly improved when solving NESs. To evaluate the proposed methods, we specifically integrate VRS with two state-of-the-art algorithms, which refer to DR-JADE (dynamic repulsion-based adaptive DE with optional external archive) \cite{liao2018solving} and MONES (the method that transforms NES to a bi-objective optimization problem) \cite{song2014locating}, respectively. Moreover, extensive experiments on two test suites, which respectively include 7 and 46 NESs, are conducted. Experimental results show that with the assistance of VRS, the methods can obtain better performance than the original methods, thus demonstrating the effectiveness of VRS for solving NESs.
	
\end{itemize}

The paper is organized as follows. Section~\ref{s2} describes NESs and briefly reviews two transformation techniques of NESs. In Section~\ref{s3}, the core idea and the reduction process of VRS are described. Then, we respectively integrate VRS with MONES and DR-JADE after presenting the framework of integrating VRS and EAs. Section~\ref{s4} selects two test suites to reduce and the relevant experiments are designed to study the superiority of the methods with VRS compared with other methods. Section~\ref{s5} concludes this paper and briefly explores future research directions.

\section{Problem description and related work}\label{s2}

\subsection{Nonlinear equations systems}\label{s21}

A NES can be formulated as:
\begin{equation}\label{Eq:NES}
f\left( {\vec x} \right) = \left[ {\begin{array}{*{20}{c}}
	{\begin{array}{*{20}{c}}
		{{f_1}\left( {\vec x} \right)}\\
		{{f_2}\left( {\vec x} \right)}
		\end{array}}\\
	{\begin{array}{*{20}{c}}
		\vdots \\
		{{f_m}\left( {\vec x} \right)}
		\end{array}}
	\end{array}} \right],
\end{equation}
where ${f_1}\left( {\vec x} \right), \cdots ,{f_m}\left( {\vec x} \right)$ indicates that the NES has $m$ functions, and at least one function is non-linear. $\vec x$ is a decision vector containing $n$ decision variables:
\begin{equation}\label{Eq:decision vector}
\vec x = \left( {{x_1},{x_2}, \cdots ,{x_n}} \right) \in {\rm{S}} \subseteq {{\rm{R}}^n},
\end{equation}
where ${\rm{S}} \subseteq {{\rm{R}}^n}$ is the decision space defined by the parametric constraints of the decision variables and it is a compact set that denotes the feasible region of the search space. The decision space can be described as:
\begin{equation}\label{Eq:decision space}
{\rm{S}} = \mathop \prod \limits_{j = 1}^n \left[ {{{\underline{x}}_j},{{\overline{x}}_j}} \right],
\end{equation}
where $j = 1, \ldots ,n$, ${\underline{x}_j}$ and ${\overline{x}_j}$ are the lower bound and upper bound of ${x_j}$.

Solving the NES shown in~(\ref{Eq:NES}) is to find a series of optimization solutions in the decision space, where each optimization solution ${\vec x^*} \in {\rm{S}}$ satisfies the following relationships:
\begin{equation}\label{Eq:solving NES}
\left\{ {\begin{array}{*{20}{c}}
	{{f_1}({{\vec x}^*}) = 0}\\
	{{f_2}({{\vec x}^*}) = 0}\\
	\vdots \\
	{{f_m}({{\vec x}^*}) = 0}
	\end{array}} \right.
\end{equation}

\begin{figure}[htb]
	\begin{center}
		\subfigure{\psfig{file=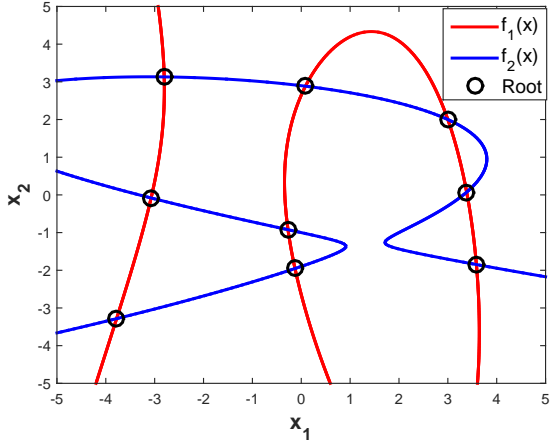,width=3.3in}}
	\end{center}
	\caption{An example of a NES problem with multiple roots}\label{Fig:example}
\end{figure}

Most of NESs have more than a single root. For instance, Fig.~\ref{Fig:example} depicts a NES problem with two nonlinear equations and two decision variables, and the expression is:
\begin{equation}\label{Eq:example NES}
\left\{ {\begin{array}{*{20}{c}}
	{4x_1^3 + 4{x_1}{x_2} + 2x_2^2 - 42{x_1} - 14 = 0} & \mbox{(a)}\\
	{4x_2^3 + 2x_1^2 + 4{x_1}{x_2} - 26{x_2} - 22 = 0} & \mbox{(b)}
	\end{array}} \right.,
\end{equation}
where ${x_i} \in [ - 5,5],{\kern 1pt} {\kern 1pt} {\kern 1pt} {\kern 1pt} {\kern 1pt} i = 1,2$.

As can be seen from Fig.~\ref{Fig:example}, the NES shown in Eq.~(\ref{Eq:example NES}) has nine roots. Each root could be equally crucial, so one of the major tasks for solving NESs is locating multiple roots. Besides, improving the quality of the found solutions is also crucial. The quality of the solutions refers to how close the solutions obtained by an algorithm is to the real solutions of a NES.

\subsection{Transformation techniques}\label{s22}

NESs are generally transformed into optimization problems to efficiently develop the roots, which possesses several advantages, such as low dependence on problem characteristics, locating multiple solutions in a single run. Currently, popular transformation techniques can be roughly classified into three categories: i) single-objective optimization-based transformation techniques \cite{liao2018solving, qin2015nonlinear, hirsch2009solving}, ii) multiobjective optimization-based transformation techniques \cite{song2014locating, qin2015nonlinear, wang2014mommop}, iii) constrained optimization-based transformation techniques \cite{mousa2008GENLS, kuri2003solution}. Herein, we will briefly introduce two representative transformation techniques about multiobjective optimization-based transformation techniques and single-objective optimization-based transformation techniques that are more commonly used transformation techniques than constrained optimization-based transformation techniques, namely Dynamic Repulsion-based EA (DREA) \cite{liao2018solving} and MONES \cite{song2014locating}, to prepare for the following research.

DREA transforms a NES to a single-objective optimization problem and locates multiple roots by repulsion techniques. The repulsive radius presented in the repulsion techniques is vital to the performance of algorithms. However, the optimal setting of the repulsive radius is arduous and problem-dependent, and it should be debugged by trial-and-error. DREA makes the repulsion radius dynamically change during the evolutionary process. The repulsion function of DREA is as follows:
\begin{equation}\label{Eq:repulsion function of DREA}
{\rm{minimize}}{\kern 1pt} {\kern 1pt} {\kern 1pt} {\kern 1pt} {\kern 1pt} {\kern 1pt} {\kern 1pt} {\kern 1pt} {\kern 1pt} {\kern 1pt} {\kern 1pt} R(\vec x) = g(\vec x) \times \prod\limits_{j = 1}^{{s_A}} {{\zeta _\gamma }(\rho ,{d_j})},
\end{equation}
where
\begin{equation}\label{Eq:objective function of DREA}
g(\vec x) = \sum\nolimits_{i = 1}^m {f_i^2(\vec x)}, 
\end{equation}
\begin{equation}\label{Eq:Euclidean distance of DREA}
{d_j} = ||\vec x - \vec x_j^*||,
\end{equation}
\begin{equation}\label{Eq:parameter of DREA}
\zeta (\rho ,{d_j}) = \left\{ {\begin{array}{*{20}{c}}
	{|{\rm{erf}}(\rho  \times {d_j}){|^{ - 1}},}&{{\rm{if}}{\kern 1pt} {\kern 1pt} {\kern 1pt} {\kern 1pt} {d_j} \le \gamma }\\
	{1,}&{{\rm{otherwise}}}
	\end{array}} \right.,
\end{equation}
where $g(\vec x)$ and $R(\vec x)$ are the objective and repulsion values of $\vec x$, respectively. ${s_A}$ is the number of roots found. ${\vec x_j^*}$ is the $j$-th root in the archive $A$ (The archive $A$ saves the roots located during the run). ${d_j}$ is the Euclidean distance of $\vec x$ and ${\vec x_j^*}$. "${\rm{erf}}$" is the error function, ${\rm{erf}}(x) = 2/\sqrt {\rm{\pi }} \int_0^x {{e^{ - {\eta ^2}}}d\eta } $. The parameter $\rho  > 0$ scales the penalty, $\rho {\rm{ = }}0.1$ is selected in the experimental section. $\gamma $ adjusts the radius of the repulsion area. The $\gamma $ of the $t$-th generation in this work is set to:
\begin{equation}\label{Eq:adjusting the radius of repulsion area}
{\gamma _t} = {\gamma _{\min }} + {\lambda _t}({\gamma _{\max }} - {\gamma _{\min }}),
\end{equation}
where
\begin{equation}\label{Eq:the min gama}
{\gamma _{\min }} = 0.01 \times \min _{i = 1}^n({\underline{x}_i} - {\overline{x}_i}),
\end{equation}
\begin{equation}\label{Eq:the max gama}
{\gamma _{\max }} = 0.5 \times \min _{i = 1}^n({\underline{x}_i} - {\overline{x}_i}),
\end{equation}
\begin{equation}\label{Eq:the parameter of gama}
{\lambda _t} = {(1 - \frac{t}{{{t_{\max }}}})^2},
\end{equation}
where $t$ is the current iteration counter. ${t_{\max }}$ is the maximal number of iterations. DREA can be classified as a multiplicative repulsion technique.

MONES transforms a NES into a bi-objective optimization problem to locate multiple roots of NESs. The transformed bi-objective optimization problem consists of two parts: the location function and the system function. The location function can be formulated as follows:
\begin{equation}\label{Eq:location function of MONES}
\left\{ {\begin{array}{*{20}{c}}
	{{\rm{minmize}}{\kern 1pt} {\kern 1pt} {\kern 1pt} {\kern 1pt} {\kern 1pt} {\alpha _1}(\vec x) = {x_1}{\kern 1pt} {\kern 1pt} {\kern 1pt} {\kern 1pt} {\kern 1pt} {\kern 1pt} {\kern 1pt} {\kern 1pt} {\kern 1pt} {\kern 1pt} {\kern 1pt} {\kern 1pt} {\kern 1pt} {\kern 1pt} {\kern 1pt} {\kern 1pt} }\\
	{{\rm{minmize}}{\kern 1pt} {\kern 1pt} {\kern 1pt} {\kern 1pt} {\kern 1pt} {\alpha _2}(\vec x) = 1 - {x_1}}
	\end{array}} \right.
\end{equation}
where ${x_1}$ is the first decision variable of the decision vector $\vec x$. Eq.~(\ref{Eq:location function of MONES}) determines the Pareto front of the optimization problem. The system function of MONES is:
\begin{equation}\label{Eq:system function of MONES}
\left\{ {\begin{array}{*{20}{c}}
	{{\rm{minmize}}{\kern 1pt} {\kern 1pt} {\kern 1pt} {\kern 1pt} {\kern 1pt} {\beta _1}(\vec x) = \sum\nolimits_{i = 1}^m {|{f_i}(\vec x)|} }\\
	{{\rm{minmize}}{\kern 1pt} {\kern 1pt} {\kern 1pt} {\kern 1pt} {\kern 1pt} {\beta _2}(\vec x) = m \times {\rm{max}}(|{f_1}(\vec x)|, \cdots ,|{f_m}(\vec x)|)}
	\end{array}} \right.
\end{equation}

Eq.~(\ref{Eq:system function of MONES}) relates the two possible transformed optimization problem versions with the NES. By combining the location function with the two optimization problem versions in Eq.~(\ref{Eq:system function of MONES}), we can derive a bi-objective optimization problem representing the original NES:
\begin{equation}\label{Eq:the transformed bi-objective optimization function of MONES}
\left\{ {\begin{array}{*{20}{c}}
	{{\rm{minmize}}{\kern 1pt} {\kern 1pt} {\kern 1pt} {\kern 1pt} {\kern 1pt} {g_1}(\vec x) = {\alpha _1}(\vec x) + {\beta _1}(\vec x){\kern 1pt} {\kern 1pt} }\\
	{{\rm{minmize}}{\kern 1pt} {\kern 1pt} {\kern 1pt} {\kern 1pt} {\kern 1pt} {g_2}(\vec x) = {\alpha _2}(\vec x) + {\beta _2}(\vec x)}
	\end{array}} \right.
\end{equation}

Eq.~(\ref{Eq:the transformed bi-objective optimization function of MONES}) makes the Pareto optimal solutions of the transformed bi-objective optimization problem correspond to the optimal solutions of the NES. Since the system functions of the NES optimization solutions are equal to 0, their images in the objective space are located on the line segment defined by $y = 1 - x$.

\section{Variable reduction strategy}\label{s3}

\subsection{Variable reduction strategy in nonlinear equation systems}\label{s31}

The main idea of VRS is to firstly explore the relationships among variables by utilizing the equality optimality condition of an optimization problem. The equality optimality condition refers to the equality condition that the optimization problem must satisfy when obtaining an optimal solution, which is expressed in the form of equations. It is a necessary condition but not necessarily a sufficient condition. For a NES, the equality optimality condition is the equations in the NES. Secondarily, according to the types and relationships of the variables, we always use a part of variables to represent and calculate the other parts of variables during the iterative search process of an EA. In this way, the variables represented and computed by other variables can be reduced and are not directly optimized (i.e., as search dimensions) during the problem-solving process. As a result, some variables and spatial dimensionality can be reduced, such that it could reduce the complexity of the problem and improve the search efficiency of the EA.

Take the NES shown in~(\ref{Eq:NES})-(\ref{Eq:solving NES}) as an example. Assume that $A$ denotes the set of decision variables included in the NES, $A{\rm{ = }}\{ {x_k}|k = 1,2, \cdots ,n\} $, ${A_j}$ is the collection of the decision variables involved in the $j$-th equation, ${A_j} \in A$. For the equation ${f_j}(\vec x){\rm{ = }}0,{\kern 1pt} {\kern 1pt} {\kern 1pt} 1 \le j \le m$, if we can obtain a relationship as:
\begin{equation}\label{Eq:reduction relationship}
{x_k} = {R_{k,j}}(\{ {x_l}|l \in {A_j},l \ne k\} )
\end{equation}

Then, in the process of locating the NES solutions, ${x_k}$ can be calculated by the relationship ${R_{k,j}}$ and the values of $\{ {x_l}|l \in {A_j},l \ne k\} $. Thus, the decision variable  ${x_k}$ can be reduced via the $j$-th equation. Meanwhile, since the variable relationship (\ref{Eq:reduction relationship}) is deduced from ${f_j}(\vec x){\rm{ = }}0$, the equation ${f_j}(\vec x){\rm{ = }}0$ is always satisfied when computing the ${x_k}$ value. Therefore, the equation ${f_j}(\vec x){\rm{ = }}0$ can be reduced as well. In addition, a constraint condition associated with the variable ${x_k}$ is added:
\begin{equation}\label{Eq:constraint condition}
{\underline{x}_k} \le {x_k} = {R_{k,j}}(\{ {x_l}|l \in {A_j},l \ne k\} ) \le {\overline{x}_k}
\end{equation}

For the sake of clear description, some key concepts are given as below:
\begin{enumerate}
	\item Core variable(s): The variable(s) used to represent other variables in the equations;
	
	\item Reduced variable(s): The variable(s) expressed and computed by core variables;
	
	\item Eliminated equation(s): The equation(s) eliminated along with the reduction of variables due to be totally satisfied by all solutions.
\end{enumerate}

For example, in Eq.~(\ref{Eq:reduction relationship}), ${x_k}$ is a reduced variable, $\{ {x_l}|l \in {A_j},l \ne k\} $ is the collection of core variables, ${f_j}(\vec x){\rm{ = }}0$ is the eliminated equation. Through the above variable reduction strategy, all variables in the NES can be divided into two categories: core variables and reduced variables. The collection of $q$ core variables is denoted as:
\begin{equation}\label{Eq:collection of core variables}
\begin{array}{l}
{X^C}{\rm{ = \{ }}{x_{c1}},{x_{c2}}, \cdots ,{x_{cq}}\} ,{\kern 1pt} {\kern 1pt} {\kern 1pt} {\kern 1pt} {\kern 1pt} {\kern 1pt} {\kern 1pt} {\kern 1pt} {\kern 1pt} q \le n\\
{{\underline{x}}_{ci}} \le {x_{ci}} \le {{\overline{x}}_{ci}},{\kern 1pt} {\kern 1pt} {\kern 1pt} {\kern 1pt} {\kern 1pt} {\kern 1pt} {\kern 1pt} {\kern 1pt} {\kern 1pt} {\kern 1pt} {\kern 1pt} {\kern 1pt} i = 1,2, \cdots ,q
\end{array}
\end{equation}

The collection composed of $l$ reduced variables is:
\begin{equation}\label{Eq:collection of reduced variables}
\begin{array}{l}
{X^R}{\rm{ = \{ }}{x_{r1}},{x_{r2}}, \cdots ,{x_{rl}}{\rm{\} }},{\kern 1pt} {\kern 1pt} {\kern 1pt} {\kern 1pt} {\kern 1pt} {\kern 1pt} {\kern 1pt} {\kern 1pt} {\kern 1pt} l{\rm{ = }}n - q\\
{{\underline{x}}_{ri}} \le {x_{ri}} \le {{\overline{x}}_{ri}},{\kern 1pt} {\kern 1pt} {\kern 1pt} {\kern 1pt} {\kern 1pt} {\kern 1pt} {\kern 1pt} {\kern 1pt} {\kern 1pt} {\kern 1pt} {\kern 1pt} {\kern 1pt} i = 1,2, \cdots ,l
\end{array}
\end{equation}

Hence, we have ${X^C} \cup {X^R} = A$ and ${X^C} \cap {X^R} = \emptyset $. The reduced decision vector can be represented by the core variables. The reduced decision space formed by the reduced decision vector is recorded as ${{\rm{S}}^*}$. The collection of $l$ eliminated equations can be denoted as:
\begin{equation}\label{Eq:collection of eliminated equations}
g = \{ {f_{s1}}(\vec x){\rm{ = }}0, \cdots ,{f_{sl}}(\vec x){\rm{ = }}0\}
\end{equation}

Accordingly, we can obtain the reduced NES:
\begin{equation}\label{Eq:part A of the reduced NES}
\left\{ {\begin{array}{*{20}{c}}
	{\begin{array}{*{20}{c}}
		{{f_1}(\vec x) = 0}\\
		{{f_2}(\vec x) = 0}
		\end{array}}\\
	{\begin{array}{*{20}{c}}
		\vdots \\
		{{f_p}(\vec x) = 0}
		\end{array}}
	\end{array}} \right.,
\end{equation}
\begin{equation}\label{Eq:part B of the reduced NES}
{\rm{s}}{\rm{.t}}{\rm{.}}{\kern 1pt} {\kern 1pt} {\kern 1pt} {\kern 1pt} {\kern 1pt} {\underline{x}_{ri}} \le {x_{ri}} = {R_{ri,sj}}({\vec x^C}) \le {\overline{x}_{ri}},{\kern 1pt} {\kern 1pt} {\kern 1pt} {\kern 1pt} {\kern 1pt} i = 1, \cdots ,l,{\kern 1pt} {\kern 1pt} {\kern 1pt} {\kern 1pt} j \in [1, \cdots ,l],
\end{equation}
\begin{equation}\label{Eq:part C of the reduced NES}
{{\rm{S}}^*} = \prod\limits_{j = 1}^q {[{{\underline{x}}_{cj}},{{\bar x}_{cj}}]},
\end{equation}
where $p$ is the number of the equations in the reduced NES, $p = m - l$. ${R_{ri,sj}}({\vec x^C})$ denotes the reduction relationship, in which the reduced variable ${x_{ri}}$ can be reduced through the $sj$-th eliminated equation.

In order to show how VRS works for a NES, consider the following quintessential illustrative example \cite{liao2018solving}:
\begin{equation}\label{Eq:a example of VRS}
\left\{ {\begin{array}{*{20}{c}}
	{3x_1^2 + \sin ({x_1}{x_2}) - x_3^2 + 2.0 = 0}\\
	{2x_1^3 + x_2^2 - {x_3} + 3.0 = 0}\\
	{\sin (2{x_1}) + \cos ({x_2}{x_3}) + {x_2} - 1.0 = 0}
	\end{array}} \right.\begin{array}{*{20}{c}}
{(a)}\\
{(b)}\\
{(c)}
\end{array},
\end{equation}
where ${x_1} \in [ - 5,5],{\kern 1pt} {\kern 1pt} {\kern 1pt} {\kern 1pt} {\kern 1pt} {x_2} \in [ - 1,3],{\kern 1pt} {\kern 1pt} {\kern 1pt} {\kern 1pt} {\kern 1pt} {x_3} \in [ - 5,5]$. A NES may have more than one reduction scheme. For~(\ref{Eq:a example of VRS}), we can choose ${x_3}$ in the equation (b) as the reduced variable, and the following reduction scheme can be obtained:
\begin{equation}\label{Eq:reduction scheme A of instance}
{x_3}{\rm{ = }}2x_1^3 + x_2^2 + 3.0
\end{equation}

Consequently, in this case, ${x_3}$ is the reduced variable. ${x_1}$ and ${x_2}$ are the core variables. The eliminated equation is $2x_1^3 + x_2^2 - {x_3} + 3.0 = 0$. The obtained reduced NES is:
\begin{equation}\label{Eq:reduced NES of instance}
\begin{array}{c}
\left\{ {\begin{array}{*{20}{c}}
	{3x_1^2 + \sin ({x_1}{x_2}) - x_3^2 + 2.0 = 0}&{(a)}\\
	{\sin (2{x_1}) + \cos ({x_2}{x_3}) + {x_2} - 1.0 = 0}&{(b)}
	\end{array}} \right.\\
{\rm{s}}{\rm{.t}}{\rm{.}}{\kern 1pt} {\kern 1pt} {\kern 1pt} {\kern 1pt} {\kern 1pt}  - 5 \le {x_3}{\rm{ = }}2x_1^3 + x_2^2 + 3.0 \le 5\\
{\kern 1pt}  - 5 \le {x_1} \le 5, - 1 \le {x_2} \le 3
\end{array}
\end{equation}

During the reduction process, constraint condition(s) associated with the reduced variable(s) should be considered. For example, reducing Eq.~(\ref{Eq:a example of VRS}) brings in the constraint conditions  presented in Eq.~(\ref{Eq:reduced NES of instance}). Furthermore, if we choose ${x_3}$ in equation (a) of Eq.~(\ref{Eq:a example of VRS}) as the reduced variable and we get:
\begin{equation}\label{Eq:reduction scheme B of instance}
{x_3} =  \pm \sqrt {3x_1^2 + \sin ({x_1}{x_2}) + 2.0}
\end{equation}

Like the reduction scheme in Eq.~(\ref{Eq:reduction scheme B of instance}), we could choose the variable with an absolute or quadratic term as a reduced variable, such that a reduced variable may have more than one value computed from the variable relationship. Therefore, the values of reduced variables that exceed the upper and lower bounds of the constraint conditions could be treated by the following handling technique.

On the one hand, just a reduced variable value corresponds to one value like Eq.~(\ref{Eq:reduction scheme A of instance}). On this occasion, when the reduced variable value is higher than its upper bound value, we make the reduced variable value equal to the upper bound value. On the contrary, we make the reduced variable value equal to the lower bound value if the reduced variable value is less than its lower bound value.

On the other hand, a reduced variable has more than one candidate value like Eq.~(\ref{Eq:reduction scheme B of instance}), we preferentially choose the value that does not violate the constraint condition. For example, when the constraint condition is $0 \le {x_1}{\rm{ = 1}} \pm {x_2} \le 1$ and the core variable ${x_2}{\rm{ = 1}}$, ${x_1}$ calculated by ${x_2}$ are either 0 or 2. ${x_1}{\rm{ = 2}}$ violates the constraint condition and is infeasible, so we make ${x_1}$ values equal to 0. If all the reduced variable values violate the constraint condition, we make them equal to its upper bound value or lower bound value.

\subsection{Integration of VRS and EA}\label{s32}

\subsubsection{A framework for integrating VRS with EA to solve NESs}

\begin{figure}[htb]
	\begin{center}
		\subfigure{\psfig{file=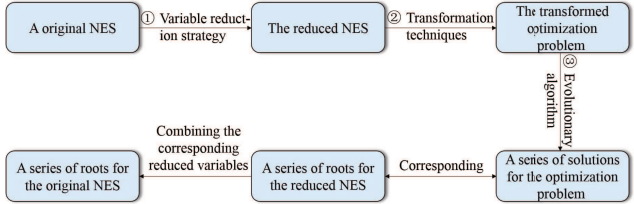,width=3.3in}}
	\end{center}
	\caption{A framework for the integration of VRS and EA}\label{Fig:framework}
\end{figure}

The framework for integrating VRS with EA is exhibited in Fig.~\ref{Fig:framework}. In this framework, a NES is first processed by VRS. The variable reduction process can be seen as a pre-processing of the NES. Second, a transformation technique is used to transform the reduced NES into an optimization problem. Then an EA is used to solve the transformed optimization problem. Thereby, a series of optimization solutions for the transformed optimization problem can be obtained, which corresponds to obtaining the roots for the reduced NES. At last, the relationships between the reduced variables and the core variables are used to compute the values of the reduced variables. By combining the values of the reduced variables and core variables, a series of roots for the original NES are finally obtained.
		

In the framework, VRS can be theoretically combined with any transformation technique and EA. In this work, we mainly study the integration of the VRS with two state-of-the-art methods, i.e., MONES and DR-JADE.

\subsubsection{The integration of VRS and DR-JADE or MONES}

MONES \cite{song2014locating} was proposed by Wang Yong et al. in 2014. MONES transforms the NES described by~(\ref{Eq:NES})-(\ref{Eq:solving NES}) into the form ~(\ref{Eq:the transformed bi-objective optimization function of MONES}), and then solves the transformed bi-objective optimization problems by NSGA-II (a fast and elitist multi-objective genetic algorithm) \cite{deb2002fast}. The method that integrates VRS into MONES is abbreviated as VR-MONES.

DREA \cite{liao2018solving} was presented by Liao Z et al. in 2020. DREA transforms a NES into the form~(\ref{Eq:repulsion function of DREA}). DREA uses JADE (adaptive differential evolution with an optional external archive) as the optimization engine \cite{zhang2009JADE}. The combined method is abbreviated as DR-JADE. We integrate VRS into DR-JADE and the resultant method is named as VR-DR-JADE for short.

\begin{figure}[htb]
	\begin{center}
		\subfigure{\psfig{file=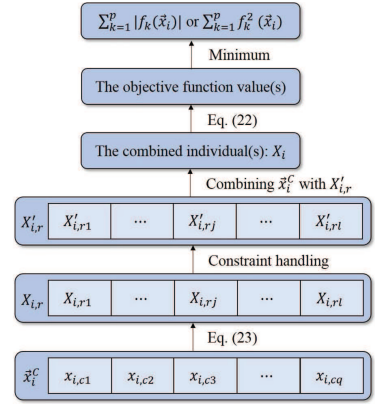,width=2.0in}}
	\end{center}
	\caption{Illustration of computing the objective function value for the individual ${\vec x_i}^C$}\label{Fig:objectivevalue}
\end{figure}

As mentioned in Section~\ref{s21}, a reduced variable may have more than one possible value, which may cause different objective function values for an individual. The objective function value refers to $\sum\nolimits_{i = 1}^p {|{f_i}(\vec x)|}$ for VR-MONES or $\sum\nolimits_{i = 1}^p {f_i^2(\vec x)}$ for VR-DR-JADE in this paper. Taking the original NES Eq.~(\ref{Eq:NES})-(\ref{Eq:solving NES}) and the reduced NES Eq.~(\ref{Eq:part A of the reduced NES})-(\ref{Eq:part C of the reduced NES}) as an example, Fig.~\ref{Fig:objectivevalue} intuitively displays the calculation process of the objective function value for an individual in a population.

In Fig.~\ref{Fig:objectivevalue}, the bottom layer is the $i$-th individual in a population during the evolution, which consists of $q$ core variable values, i.e., ${\vec x_i}^C{\rm{ = (}}{x_{i,c1}},{x_{i,c2}}, \cdots ,{x_{i,cq}})$. First, we compute the reduced variable values via the core variable values in ${\vec x_i}^C$ and the variable relationships expressed in Eq.~(\ref{Eq:part B of the reduced NES}). The obtained reduced variable values are put in the set ${X_{i,r}}$, where ${X_{i,rj}}$ denotes the set of the value(s) of the $j$-th reduced variable (a reduced variable may have more than one candidate values). Second, we handle the reduced variable values violating the constraint by the technique introduced in Section~\ref{s21} and thus we obtain the feasible reduced variables set $X{'_{i,r}}$. Then, we combine the values of the reduced variables in the set $X{'_{i,r}}$ with the individual ${\vec x_i}^C$ (i.e., the set of core variable values) to form the new individual(s) denoted by ${X_i}$. After that, values in ${X_i}$ is substituted into the reduced equations~(\ref{Eq:part A of the reduced NES}) to compute the objective function value(s). Finally, we select the minimum value of $\sum\nolimits_{k = 1}^p {|{f_k}(\vec x)|}$ or $\sum\nolimits_{k = 1}^p {f_k^2(\vec x)}$ as the objective function value of the individual ${\vec x_i}^C$.

Next, for VR-MONES, we should compute the transformed bi-objective optimization function by Eq.~(\ref{Eq:location function of MONES})-(\ref{Eq:the transformed bi-objective optimization function of MONES}), in which ${x_1}$ is the first decision variable of the decision vector ${\vec x^C}$ after reduction. For VR-DR-JADE, the objective function value obtained by Fig.~\ref{Fig:objectivevalue} is the $g(\vec x)$ value of the individual ${\vec x_i}^C$. We can compute the value of the corresponding $R(\vec x)$ by Eq.~(\ref{Eq:repulsion function of DREA}). Then, we can use VR-MONES or VR-DR-JADE to iterate and get an evolved population $pop$. Finally, combining the values of core variable in $pop$ and the values of reduced variable computed by reduction relationships forms the final population.

\section{Experimental study}\label{s4}

To demonstrate that VRS can improve the performance of the original algorithms, this section mainly focuses on the comparison between VR-MONES and VR-DR-JADE with their corresponding original algorithms, i.e., MONES and DR-JADE, respectively. In Section~\ref{s41}, we use a benchmark suite with 7 NESs (in which two test problems are real-world problems) to test the efficiency of VRS by comparing the experimental results between VR-MONES and MONES. Moreover, in Section~\ref{s42}, a large scale test suite of 46 NESs (in which five test problems are real-world problems) is used to testify the effectiveness of VRS by comparing VR-DR-JADE with other popular and state-of-the-art methods. In Section~\ref{s43}, we briefly conclude the experimental results obtained by Section~\ref{s41} and Section~\ref{s42}.

\subsection{Experimental study on VR-MONES}\label{s41}

We perform VRS for the 7 NESs in reference \cite{song2014locating}, and compare the performance of VR-MONES and MONES in terms of two performance indicators, i.e., the inverted generational distance and the number of the optimal solutions found.

\subsubsection{Test problems}

In this section, seven test problems (denoted as F1-F7) are used to investigate the effectiveness of VR-MONES. Among them, the optimal solutions of F1-F4 are known. F5-F7 have infinitely optimal solutions, which are not completely known up to now. F6 and F7 are real-world problems and are derived from neurophysiology application model and economics system model. The brief information on the seven test problems is summarized in Table~\ref{Table:information of datasets F1-F7}, including the number of the decision variables ($D$), the decision space ($S$), the number of the linear equations ($LE$), the number of the nonlinear equations ($NE$), and the number of the roots ($NoR$).

\aboverulesep=0pt \belowrulesep=0pt
\newcommand{\tabincell}[2]{\begin{tabular}{@{}#1@{}}#2\end{tabular}}
\begin{table}[htp]
	\footnotesize
	\renewcommand{\arraystretch}{1.1}
	\caption{Characterizations of the test problems F1-F7.}
	\label{Table:information of datasets F1-F7}
	\begin{center}
		\begin{tabular}{cccccc} \toprule
			
			Test instance & $D$ & $S$ & $LE$ & $NE$ & $NoR$ \\ 
			\midrule
			
			F1 & 2 & ${[ - 1,1]^2}$ & 1 & 1 & 2 \\
			
			F2 & 20 & ${[ - 1,1]^{20}}$ & 0 & 2 & 2 \\
			
			F3 & 2 & ${[ - 1,1]^2}$ & 1 & 1 & 11 \\
			
			F4 & 2 & ${[ - 1,1]^2}$ & 0 & 2 & 15 \\
			
			F5 & 3 & ${[ - 1,1]^3}$ & 1 & 1 & infinite \\
			
			F6 & 6 & ${[ - 1,1]^6}$ & 0 & 6 & infinite \\
			
			F7 & 20 & ${[ - 1,1]^{20}}$ & 1 & 19 & infinite \\
			
			\bottomrule
			
		\end{tabular}
	\end{center}
\end{table}

\subsubsection{Performance metrics}

In this section, two performance indicators are introduced to evaluate the capability of VR-MONES and MONES to locate the roots of a NES.

\begin{enumerate}
	\item The inverted generational distance (IGD) \cite{wang2012regularity}: The IGD indicator is computed as:
	\begin{equation}\label{Eq:IGD}
	{\rm{IGD}}(IP,I{P^*}) = \frac{{\sum\nolimits_{i = 1}^{|I{P^*}|} {d({{\vec \nu }_i},IP)} }}{{|I{P^*}|}},
	\end{equation}
	where $IP$ is a set of the images of the individuals of a population in the objective space and $I{P^*}$ is a set of the images of all optimal solutions of a NES in the objective space: $I{P^*}{\rm{ = }}\{ {\vec \nu _1}, \cdots ,{\vec \nu _2}\} $. $d({\vec \nu _i},IP)$ is the minimum Euclidean distance between ${\vec \nu _i}$ and the points in $IP$. If a NES (such as F5, F6 or F7) has infinite roots, $|I{P^*}|$ is a set of uniformly distributed points in the objective space along Pareto front. $|I{P^*}|$ is the number of optimal solutions in $I{P^*}$, we set ${I{P^*} = 100}$ for F5, F6 and F7. In this section, the objective space is defined by $x = {x_r}$ and $y = 1 - {x_r}$ for MONES or VR-MONES. ${x_r}$ is the first decision variable of a NES for MONES or the first decision variable of a reduced NES for VR-MONES.

	IGD can measure both the diversity and convergence of $IP$.

	\item Number of the optimal solutions found (NOF) \cite{song2014locating}: The NOF indicator is computed as:
	\begin{equation}\label{Eq:NOF}
	{\rm{NOF}}(IP,I{P^*}) = \sum\nolimits_{i = 1}^{|I{P^*}|} {flag({{\vec \nu }_i})},
	\end{equation}
	where
	\begin{equation}\label{Eq:flag of NOF}
	\left\{ {\begin{array}{*{20}{c}}
		{flag({{\vec \nu }_i}) = 1,}&{{\rm{if}}{\kern 1pt} {\kern 1pt} {\kern 1pt} {\kern 1pt} d({{\vec \nu }_i},IP) \le \varepsilon ,{{\vec \nu }_i} \in I{P^{\rm{*}}}}\\
		{flag({{\vec \nu }_i}) = 0,}&{{\rm{otherwise}}{\kern 1pt} {\kern 1pt} {\kern 1pt} {\kern 1pt} {\kern 1pt} {\kern 1pt} {\kern 1pt} {\kern 1pt} {\kern 1pt} {\kern 1pt} {\kern 1pt} {\kern 1pt} {\kern 1pt} {\kern 1pt} {\kern 1pt} {\kern 1pt} {\kern 1pt} {\kern 1pt} {\kern 1pt} {\kern 1pt} {\kern 1pt} {\kern 1pt} {\kern 1pt} {\kern 1pt} {\kern 1pt} {\kern 1pt} {\kern 1pt} {\kern 1pt} {\kern 1pt} {\kern 1pt} {\kern 1pt} {\kern 1pt} {\kern 1pt} {\kern 1pt} {\kern 1pt} {\kern 1pt} {\kern 1pt} {\kern 1pt} {\kern 1pt} {\kern 1pt} {\kern 1pt} {\kern 1pt} {\kern 1pt} {\kern 1pt} {\kern 1pt} {\kern 1pt} {\kern 1pt} {\kern 1pt} {\kern 1pt} {\kern 1pt} {\kern 1pt} {\kern 1pt} }
		\end{array}} \right.
	\end{equation}
	
	Here, $\varepsilon $ is a user-defined threshold value. In this section, we set $\varepsilon {\rm{ = }}0.01$ for F5 and 0.02 for the other six problems according to the number of decision variables \cite{song2014locating}. The larger the NOF-indicator value is, the more roots are found.
\end{enumerate}

\subsubsection{Variable reduction results}

According to the reduction method given in Section~\ref{s21}, a NES may have more than one reduction scheme. In this section, we only show one reduction scheme thought to be promising for each NES. The expressions of the 7 NESs and the related reduction schemes are shown in Table~\ref{Table:variable reduction of datasets F1-F7}. The successive experiments are based on the reduction schemes shown in Table~\ref{Table:variable reduction of datasets F1-F7}.

\aboverulesep=0pt \belowrulesep=0pt
\begin{table*}[htp]
	
	\footnotesize
	\renewcommand{\arraystretch}{0.9}
	\centering
	\caption{Expressions and variable reduction of test suite F1-F7.}
	\label{Table:variable reduction of datasets F1-F7}
	\begin{center}
		\begin{tabular}{ccc} \toprule
			Test instance & The expression of NES & Reduced variable(s) and eliminated equation(s) \\ \midrule
			
			\multirow{2}{*}{F1} & $x_1^2 + x_2^2 - 1 = 0{\kern 1pt} {\kern 1pt} {\kern 1pt} {\kern 1pt} {\kern 1pt} (1)$ & Equation (2) \\
			& ${x_1} - {x_2} = 0{\kern 1pt} {\kern 1pt} {\kern 1pt} {\kern 1pt} {\kern 1pt} (2)$ & ${x_2} = {x_1}$ \\ \hline

			\multirow{2}{*}{F2} & $\sum\nolimits_{i = 1}^D {x_i^2}  - 1 = 0{\kern 1pt} {\kern 1pt} {\kern 1pt} {\kern 1pt} {\kern 1pt} (1)$ & Equation (1) \\
			& $|{x_1} - {x_2}| + \sum\nolimits_{i = 3}^D {x_i^2}  = 0{\kern 1pt} {\kern 1pt} {\kern 1pt} {\kern 1pt} {\kern 1pt} (2)$ & ${x_2} =  \pm \sqrt {1 - (x_1^2 + \sum\nolimits_{i = 3}^D {x_i^2} )} $ \\ \hline

			\multirow{2}{*}{F3} & ${x_1} - \sin (5\pi x) = 0{\kern 1pt} {\kern 1pt} {\kern 1pt} {\kern 1pt} {\kern 1pt} (1)$ & Equation (2) \\
			& ${x_1} - {x_2} = 0{\kern 1pt} {\kern 1pt} {\kern 1pt} {\kern 1pt} {\kern 1pt} (2)$ & ${x_2} = {x_1}$ \\ \hline

			\multirow{2}{*}{F4} & ${x_1} - \cos (4\pi {x_2}) = 0{\kern 1pt} {\kern 1pt} {\kern 1pt} {\kern 1pt} {\kern 1pt} (1)$ & Equation (1) \\
			& $x_1^2 + x_2^2 = 1{\kern 1pt} {\kern 1pt} {\kern 1pt} {\kern 1pt} {\kern 1pt} (2)$ & ${x_1}{\rm{ = }}\cos (4\pi {x_2})$ \\ \hline
			
			F5 & \tabincell{c}{${x_1} + {x_2} + {x_3} - 1 = 0{\kern 1pt} {\kern 1pt} {\kern 1pt} {\kern 1pt} {\kern 1pt} (1)$ \\ ${x_1} - x_2^3 = 0{\kern 1pt} {\kern 1pt} {\kern 1pt} {\kern 1pt} {\kern 1pt} (2)$} & \tabincell{c}{Equation (1) and (2) \\ ${x_1}{\rm{ = }}x_2^3$ \\ ${x_3} = 1 - {x_1} - {x_2}$} \\ \hline

			F6 & \tabincell{c}{$x_1^2 + x_3^2 = 1{\kern 1pt} {\kern 1pt} {\kern 1pt} {\kern 1pt} {\kern 1pt} (1)$ \\ $x_2^2 + x_4^2 = 1{\kern 1pt} {\kern 1pt} {\kern 1pt} {\kern 1pt} {\kern 1pt} (2)$ \\ ${x_5}x_3^3 + {x_6}x_4^3 = 0{\kern 1pt} {\kern 1pt} {\kern 1pt} {\kern 1pt} {\kern 1pt} (3)$ \\ ${x_5}x_1^3 + {x_6}x_2^3 = 0{\kern 1pt} {\kern 1pt} {\kern 1pt} {\kern 1pt} {\kern 1pt} (4)$ \\ ${x_5}{x_1}x_3^2 + {x_6}x_4^2{x_2} = 0{\kern 1pt} {\kern 1pt} {\kern 1pt} {\kern 1pt} {\kern 1pt} (5)$ \\ ${x_5}{x_3}x_1^2 + {x_6}x_2^2{x_4} = 0{\kern 1pt} {\kern 1pt} {\kern 1pt} {\kern 1pt} {\kern 1pt} (6)$} & \tabincell{c}{Equation (1), (2) and (3) \\ ${x_1} =  \pm \sqrt {1 - x_3^2} $ \\ ${x_2} =  \pm \sqrt {1 - x_4^2} $ \\ ${x_6} =  - {x_5}x_3^3/x_4^3$} \\ \hline

			F7 & \tabincell{c}{$({x_k} + \sum\nolimits_{i = 1}^{D - k - 1} {{x_i}{x_{i + k}}){x_D} - {{\rm{c}}_k}}  = 0,$\\ $1 \le k \le D - 1{\kern 1pt} {\kern 1pt} {\kern 1pt} {\kern 1pt} {\kern 1pt} (1)$ \\$\sum\nolimits_{l = 1}^{D - 1} {{x_l} + 1 = 0} {\kern 1pt} {\kern 1pt} {\kern 1pt} {\kern 1pt} {\kern 1pt} (2)$} & \tabincell{c}{Equation (2) \\ ${x_{19}} = 1 - \sum\nolimits_{j = 1}^{18} {{x_j}} $} \\	
			\bottomrule	
			
		\end{tabular}
	\end{center}
\end{table*}

As can be seen from Table~\ref{Table:variable reduction of datasets F1-F7}, each problem in F1-F4 contains two decision variables and two equations, in which one variable and one equation can be reduced. Two variables and all the two equations can be reduced for NES F5. Three variables and three equations can be reduced for NES F6, and the reduced F6 contains three variables and equations. In this Section, we set ${c_k} = 0,1 \le k \le D - 1$ for F7. One variable and equation can be reduced for NES F7.

\subsubsection{Experimental results and discussions on F1-F7}

For a fair comparison, the parameter settings of VR-MONES are the same as those of the original MONES in \cite{song2014locating}. To make the experimental results reliable, 30 independent runs are executed on each NES, and the maximum number of generations is set to 500 (i.e., the maximum function evaluation number is 50,000) for each run. Table~\ref{Table:results of VR-MONES} presents the best, mean, worst and standard deviation of the IGD-indictor and NOF-indicator values generated by VR-MONES and MONES.

\aboverulesep=0pt \belowrulesep=0pt
\begin{table}[htp]
	\footnotesize
	\centering
	\caption{Status of IGD-indicator and NOF-indicator in VR-MONES and MONES. The better or equal IGD-indicator and NOF-indicator for each NES are highlighted in boldface.}
	\label{Table:results of VR-MONES}
	\begin{center}
		\begin{tabular}{cccccc} \toprule
			\multirow{2}{*}{\tabincell{c}{Test \\instance}} & \multirow{2}{*}{Status} & \multicolumn{2}{c}{IGD} & \multicolumn{2}{c}{NOF} \\
			\cline{3-6}
			&   & MONES & \tabincell{c}{VR- \\MONES} & MONES & \tabincell{c}{VR- \\MONES} \\
			\midrule

			\multirow{4}{*}{F1} & Best & 1.51E-04 & {\bf 1.47E-04} & 2.00E+00 & {\bf 2.00E+00} \\
			& Mean & 2.05E-04 & {\bf 1.57E-04} & 2.00E+00 & {\bf 2.00E+00} \\
			& Worst & 4.00E-04 & {\bf 2.14E-04} & 2.00E+00 & {\bf 2.00E+00} \\
			& Std & 6.92E-05 & {\bf 1.34E-05} & 0.00E+00 & {\bf 0.00E+00} \\	 \hline

			\multirow{4}{*}{F2} & Best & 1.95E-04 & {\bf 1.48E-04} & 2.00E+00 & {\bf 2.00E+00} \\
			& Mean & 6.09E-04 & {\bf 1.71E-04} & 2.00E+00 & {\bf 2.00E+00} \\
			& Worst & 1.46E-03 & {\bf 2.68E-04} & 2.00E+00 & {\bf 2.00E+00} \\
			& Std & 2.85E-04 & {\bf 2.74E-05} & 0.00E+00 & {\bf 0.00E+00} \\	 \hline

			\multirow{4}{*}{F3} & Best & 1.34E-03 & {\bf 9.90E-05} & 1.10E+01 & {\bf 1.10E+01} \\
			& Mean & 3.82E-03 & {\bf 1.78E-04} & 1.09E+01 & {\bf 1.10E+01} \\
			& Worst & 2.63E-02 & {\bf 3.06E-04} & 1.00E+01 & {\bf 1.10E+01} \\
			& Std & 6.02E-03 & {\bf 4.76E-05} & 2.54E-01 & {\bf 0.00E+00} \\	 \hline

			\multirow{4}{*}{F4} & Best & 2.75E-03 & {\bf 2.05E-03} & 1.50E+01 & {\bf 1.50E+01} \\
			& Mean & 1.23E-02 & {\bf 2.20E-03} & 1.42E+01 & {\bf 1.50E+01} \\
			& Worst & 9.00E-02 & {\bf 2.63E-03} & 1.10E+01 & {\bf 1.50E+01} \\
			& Std & 1.73E-02 & {\bf 1.08E-04} & 1.15E+00 & {\bf 0.00E+00} \\	 \hline

			\multirow{4}{*}{F5} & Best & 1.56E-02 & {\bf 5.36E-03} & 5.00E+01 & {\bf 9.00E+01} \\
			& Mean & 4.08E-02 & {\bf 5.96E-03} & 3.74E+01 & {\bf 8.35E+01} \\
			& Worst & 1.78E-01 & {\bf 6.63E-03} & 1.90E+01 & {\bf 7.60E+01} \\
			& Std & 3.82E-02 & {\bf 3.77E-04} & 7.25E+00 & {\bf 3.10E+00} \\	 \hline

			\multirow{4}{*}{F6} & Best & 1.31E-02 & {\bf 8.83E-03} & 8.50E+01 & {\bf 9.50E+01} \\
			& Mean & 2.21E-02 & {\bf 1.10E-02} & 7.33E+01 & {\bf 8.86E+01} \\
			& Worst & 4.94E-02 & {\bf 1.18E-02} & 6.00E+01 & {\bf 8.40E+01} \\
			& Std & 8.93E-03 & {\bf 7.59E-04} & 5.33E+00 & {\bf 2.86E+00} \\	 \hline

			\multirow{4}{*}{F7} & Best & 4.09E-02 & {\bf 8.13E-03} & 3.60E+01 & {\bf 9.80E+01} \\
			& Mean & 1.84E-01 & {\bf 9.44E-03} & 1.43E+01 & {\bf 9.14E+01} \\
			& Worst & 8.92E-01 & {\bf 1.06E-02} & 0.00E+00 & {\bf 8.60E+01} \\
			& Std & 2.24E-01 & {\bf 6.16E-04} & 1.07E+01 & {\bf 2.59E+00} \\	 
			\bottomrule

		\end{tabular}
	\end{center}
\end{table}
From the results in Table~\ref{Table:results of VR-MONES}, in regard to the IGD indicator, the best, mean, worst and standard deviation of the IGD-indicator values obtained by VR-MONES have significantly improved for all the test problems. For example, for NES F3, the best, mean, worst and standard deviation of the IGD-indicator values obtained by VR-MONES have been respectively improved by 92.61$\%$, 95.34$\%$, 98.84$\%$ and 99.21$\%$ compared with those obtained by MONES. The phenomenon indicates that the solutions obtained by VR-MONE are closer to the actual known solutions for all the test problems than those obtained by MONES. We also implemented the Wilcoxon test on the mean IGD-indicator for all the test problem over 30 runs\footnote{The statistical tests reported in this paper are caculated by the KEEL3.0 software \cite{triguero2017keel}.}. Compared VR-MONES with MONES, we can get ${R^ + } = 28.0$, ${R^ - } = 0.0$ and $p = {\rm{1}}{\rm{.56E - 02}}$ by the Wilcoxon test. Since VR-MONES can provide higher ${R^ + }$ value than ${R^ - }$ value and the $p$ value is less than 0.05, VR-MONES is significantly better than MONES on the seven test problems.

With respect to the NOF-indicator, it is clear that the best, mean, worst and standard deviation of NOF-indicator values obtained by VR-MONES are better or at least equal to those obtained by MONES for NESs F1-F7. The results reveal that VR-MONES can find more roots than MONES. For each NES in F1-F4 with known optimal solutions, VR-MONES can successfully locate all the roots over 30 runs. For each NES in F5-F7 with infinitely many roots (the default number of the optimal solutions is 100 in this section), VR-MONES has the capability to maintain much more roots than MONES, especially for F5 and F7.

\begin{figure*}[htp]
	\begin{center}
		\subfigure[\small{F1}]{\psfig{file=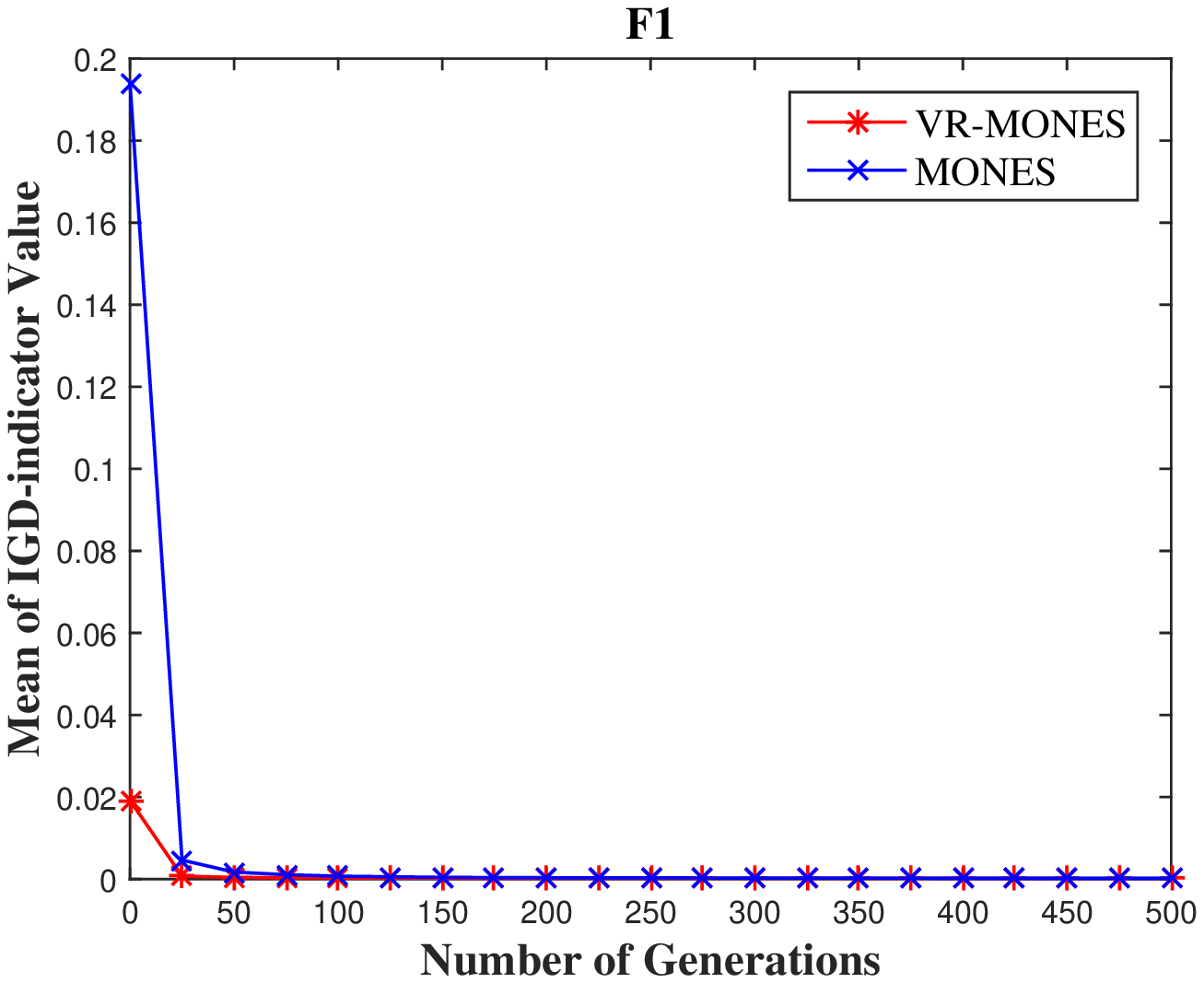,width=1.55in}}
		\subfigure[\small{F2}]{\psfig{file=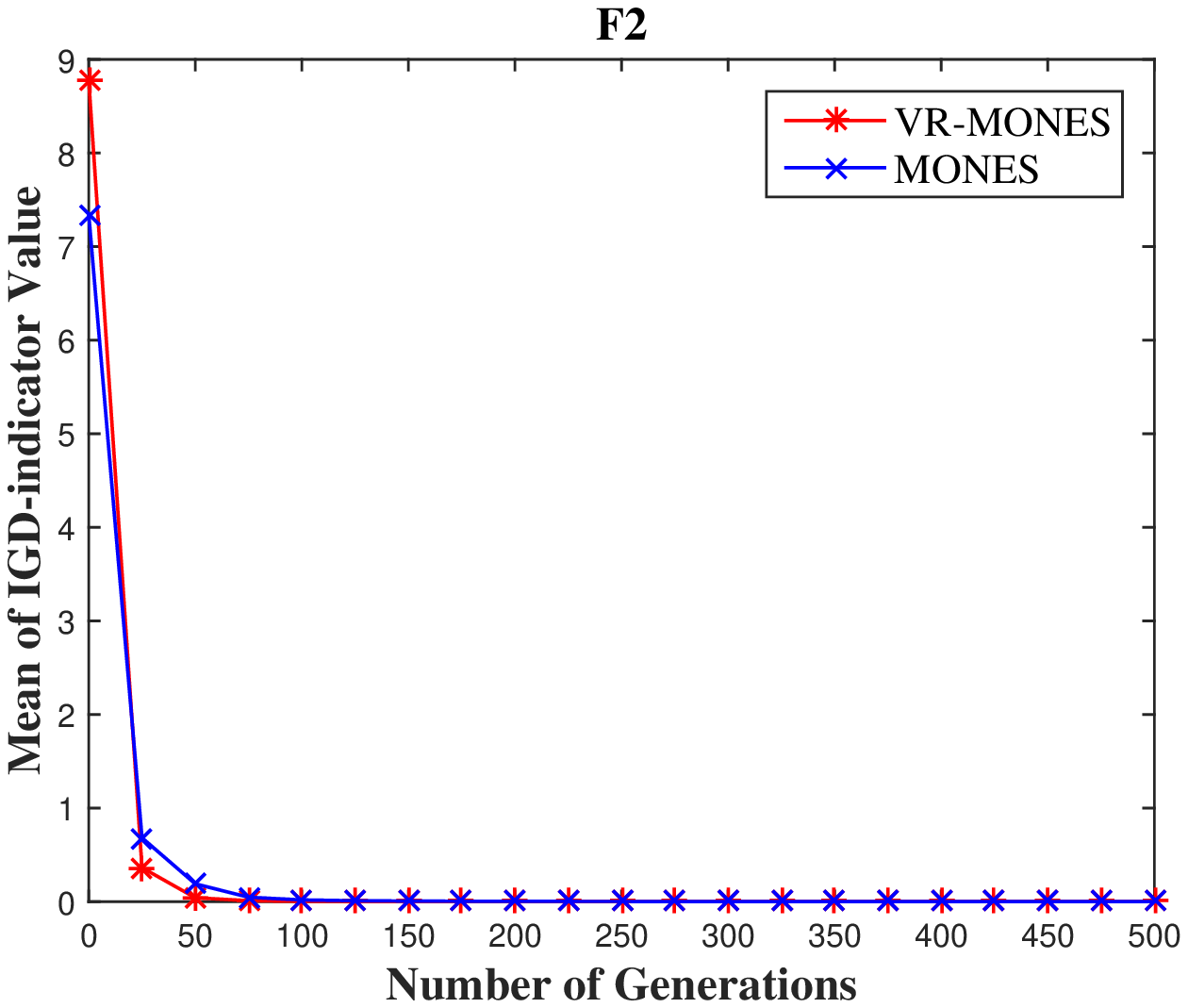,width=1.55in}}
		\subfigure[\small{F3}]{\psfig{file=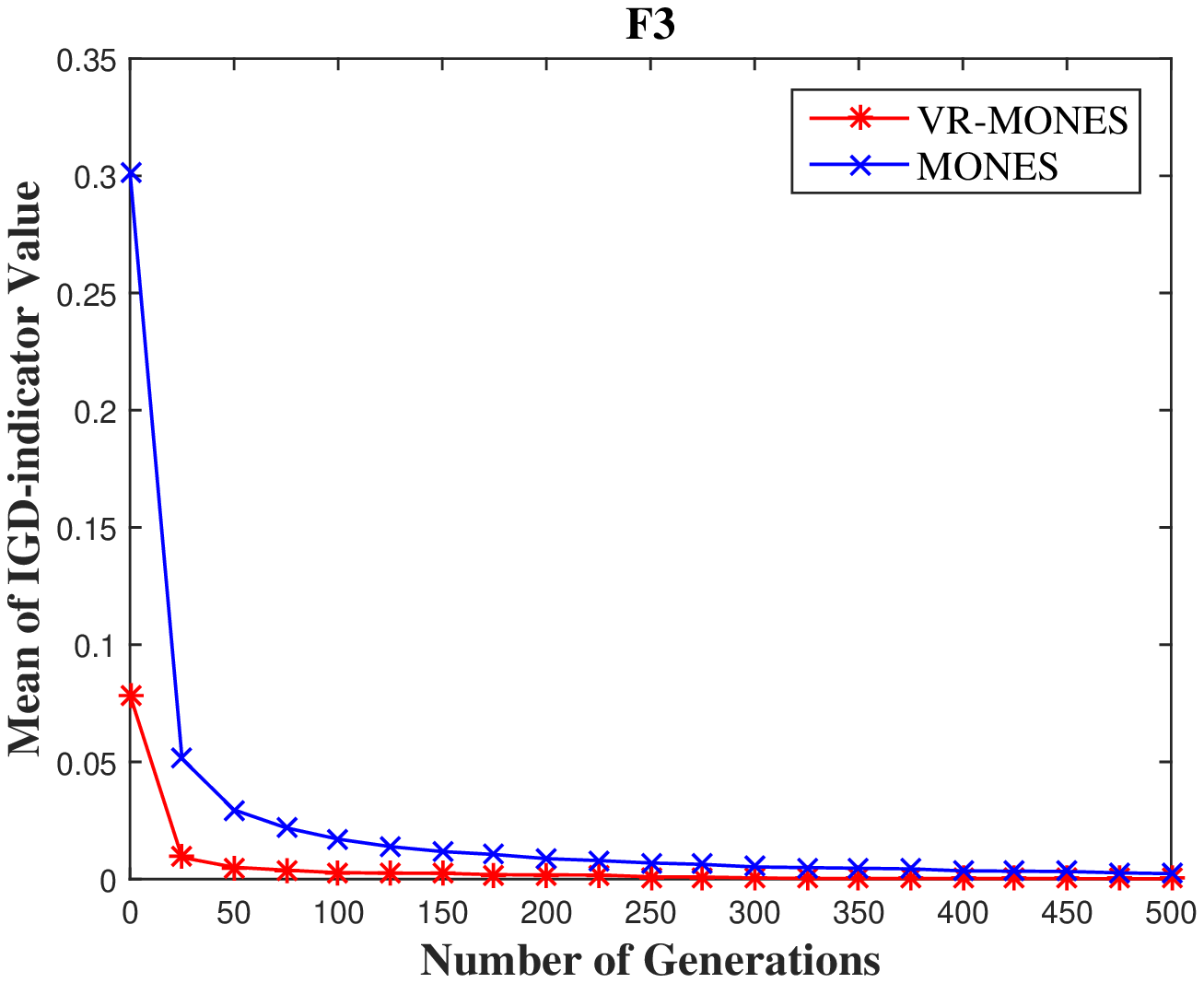,width=1.55in}}
		\subfigure[\small{F4}]{\psfig{file=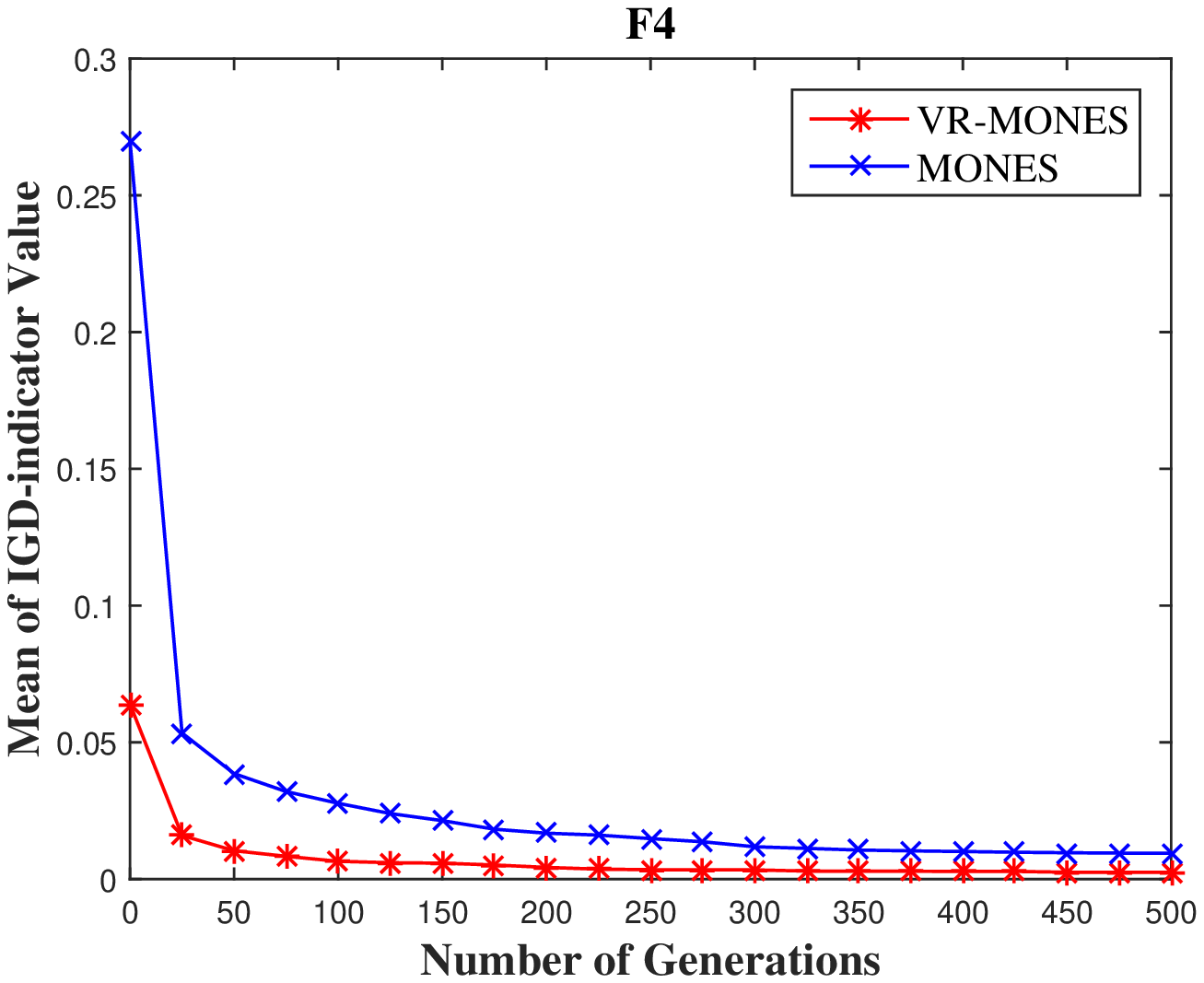,width=1.55in}}
		\subfigure[\small{F5}]{\psfig{file=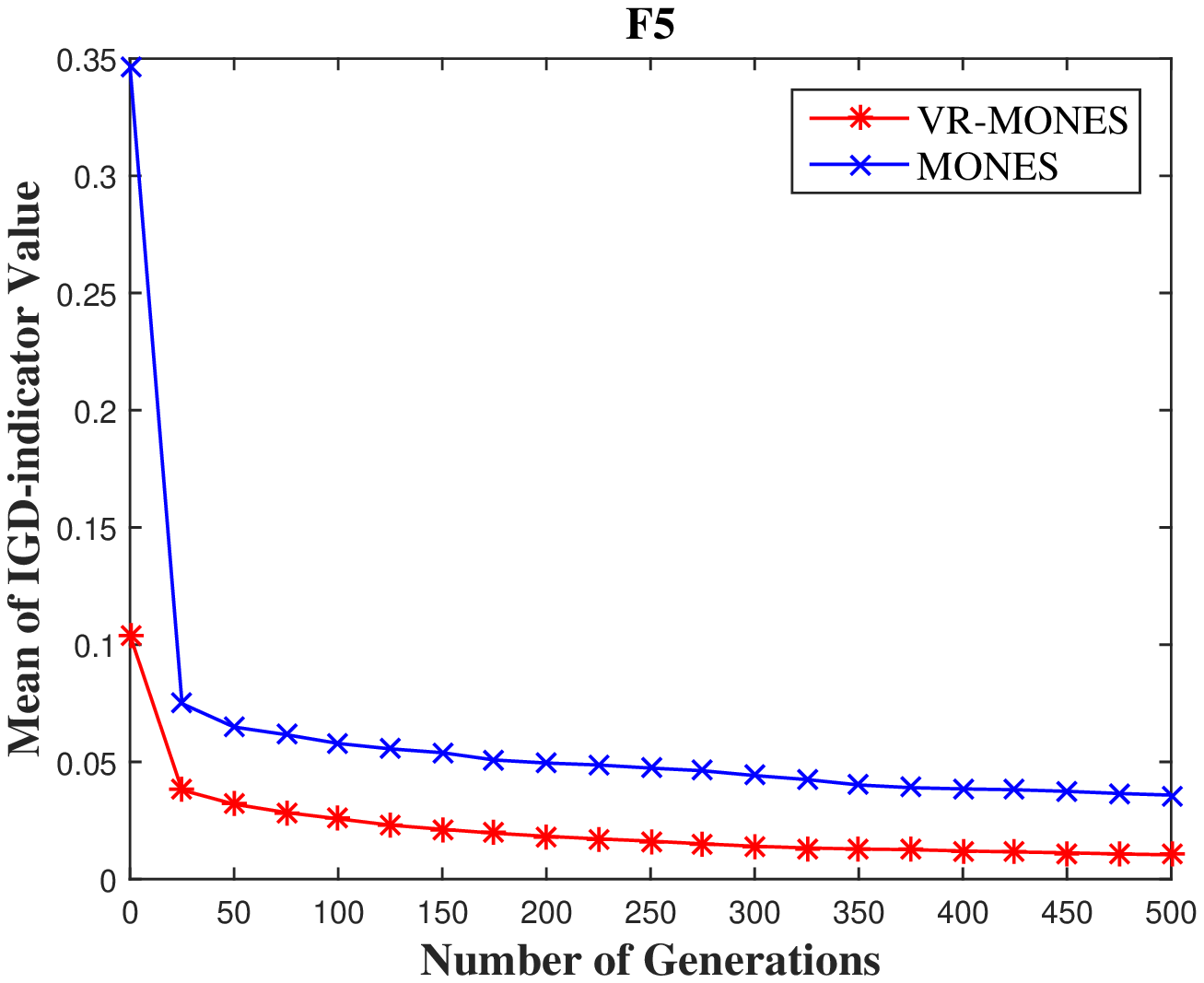,width=1.55in}}
		\subfigure[\small{F6}]{\psfig{file=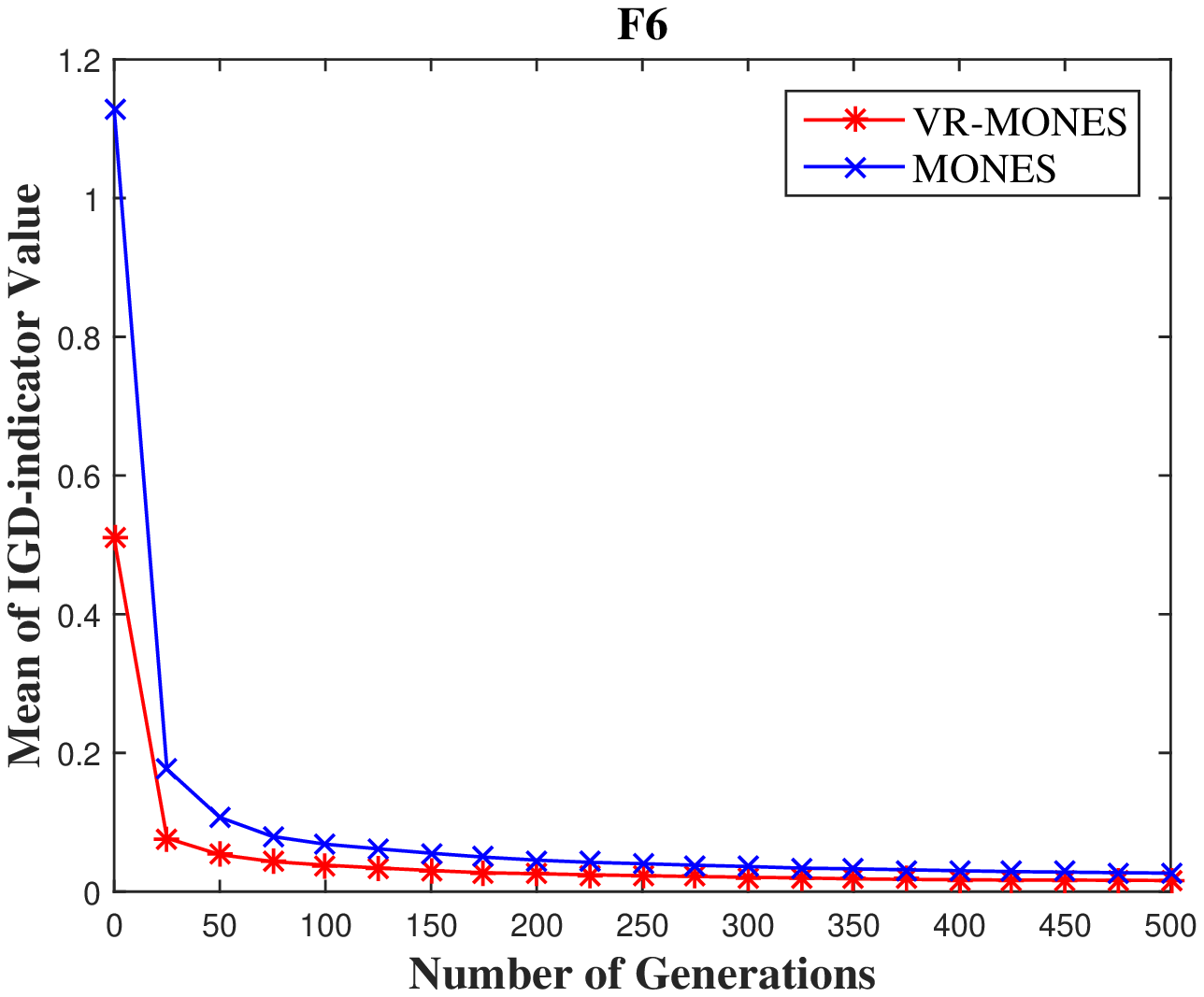,width=1.55in}}
		\subfigure[\small{F7}]{\psfig{file=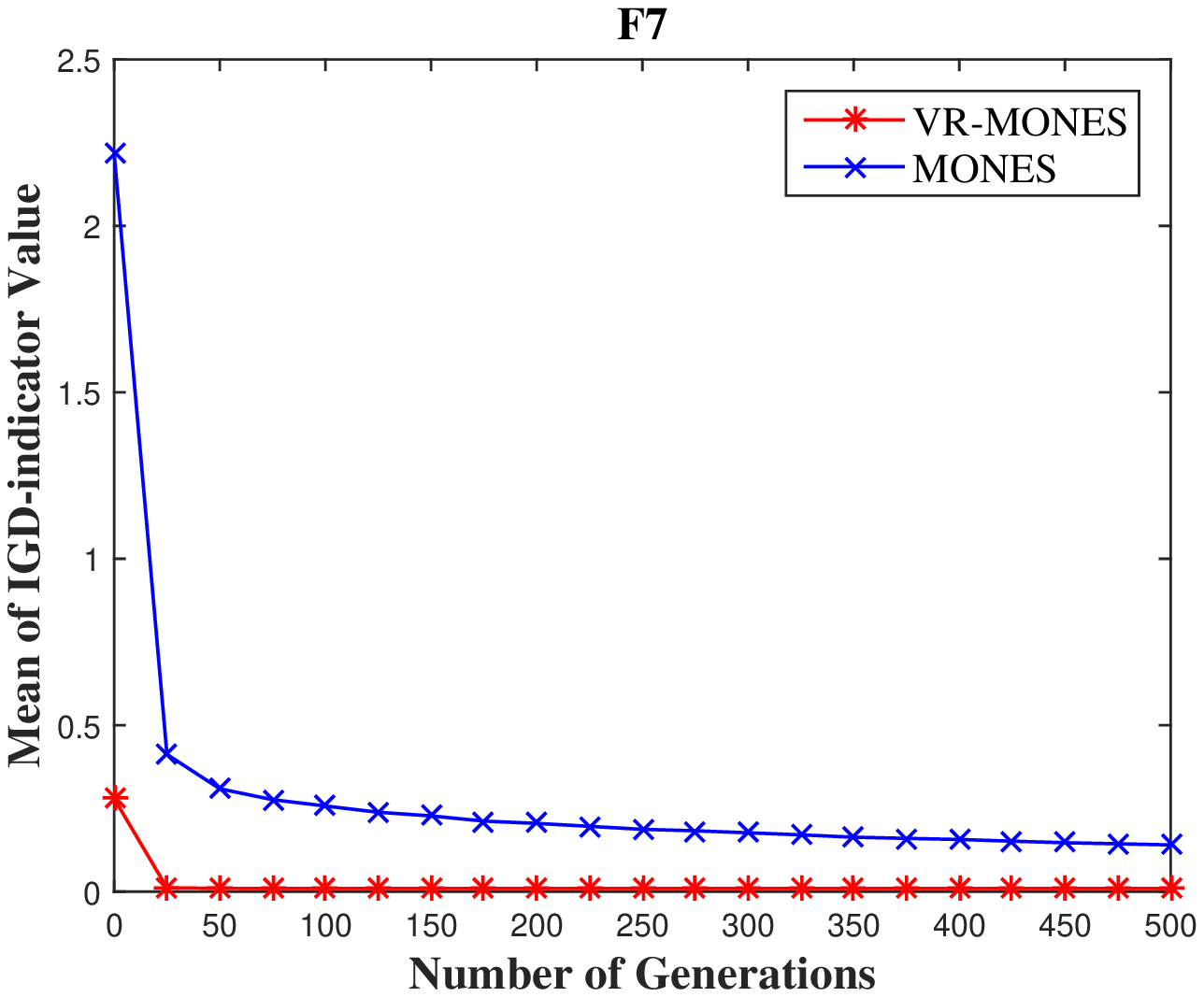,width=1.55in}}
	\end{center}\vspace{-3mm}
	\caption{Mean of IGD-indicator values for VR-MONES and MONES during the evolution} \label{Fig:convergence curve}
\end{figure*}

To further show the performance of MONES and VR-MONES, Fig.~\ref{Fig:convergence curve} provides the convergence process of the mean IGD-indicator values provided by MONES and VR-MONES for all the test problems over 30 independent runs.

As depicted in Fig.~\ref{Fig:convergence curve}, at the beginning of evolution, the convergence curve of VR-MONES or MONES starts with a relatively large mean IGD-indicator value. As the optimization proceeding, the mean IGD-indicator values in VR-MONES and MONES both converge continuously toward a positive number close to 0. Particularly, VR-MONES can converge to a smaller IGD-indicator value for each NES in F1-F7. For example, during the evolution, the mean IGD-indicator values of MONES and VR-MONES start at 0.2699 and 0.064 and eventually converge to 0.0094 and 0.0025 respectively for NES F4. This reveals that VR-MONES can robustly obtain better solutions while maintaining the diversity of the population and that the integration of VRS can noticeably improve the search efficiency of MONES.

\subsection{Experimental study on VR-DR-JADE}\label{s42}

To further evaluate the effectiveness of VRS, this section applies VRS to another test suite with 46 NESs. In this case, the representative algorithm DR-JADE with and without VRS are used to solve the test problems, respectively. The performance of VR-DR-JADE and other state-of-the-art methods is evaluated by the values of root ratio, success rate and other indicators obtained from the experiments.

\subsubsection{Test problems}

In this section, we choose 46 NESs with diverse features to extensively evaluate the performance of an algorithm. The brief information and the maximal number of function evaluations ($NFE{s_{\max }}$) of the 46 NESs are shown in Table~\ref{Table:information of datasets E1-E46}, in which the  meaning of "\emph{D}", "\emph{LE}", "\emph{NE}", and "\emph{NoR}" are the same as Table~\ref{Table:information of datasets F1-F7}. The optimal solutions of NESs E1-E42 are known, while the optimal solutions of NESs F43-F46 are unknown. NESs E13 and E43-E46 come from real-world applications. $NFE{s_{\max }}$ values are different for different problems owing to their different difficulties. In this section, to fairly compare the performance of VR-DR-JADE and DR-JADE, we use the same $NFE{s_{\max }}$ as literature \cite{liao2018solving}.

\aboverulesep=0pt \belowrulesep=0pt
\begin{table}[htp]
	\footnotesize

	\caption{Brief description and the $NFE{s_{\max }}$ of test problems E1-E46.}
	\label{Table:information of datasets E1-E46}
	\begin{center}
		\begin{tabular}{cccccc} \toprule
			Test instance & \emph{D} & \emph{LE} & \emph{NE} & \emph{NoR} & $NFE{s_{\max }}$ \\
			\bottomrule

			E1 & 2 & 0 & 2 & 2 & 10 000 \\
			
			E2 & 2 & 0 & 2 & 3 & 50 000 \\
			
			E3 & 3 & 0 & 3 & 2 & 100 000 \\
			
			E4 & 2 & 1 & 1 & 2 & 20 000 \\
			
			E5 & 2 & 0 & 2 & 3 & 100 000 \\
			
			E6 & 3 & 1 & 2 & 3 & 50 000 \\
			
			E7 & 2 & 0 & 2 & 3 & 50 000 \\
			
			E8 & 2 & 1 & 1 & 2 & 50 000 \\
			
			E9 & 20 & 0 & 2 & 2 & 50 000 \\
			
			E10 & 2 & 1 & 1 & 11 & 50 000 \\
			
			E11 & 2 & 0 & 2 & 15 & 50 000 \\
			
			E12 & 2 & 0 & 2 & 13 & 50 000 \\
			
			E13 & 10 & 0 & 10 & 1 & 50 000 \\
			
			E14 & 2 & 1 & 1 & 8 & 50 000 \\
			
			E15 & 4 & 0 & 4 & 1 & 50 000 \\
			
			E16 & 2 & 0 & 2 & 7 & 50 000 \\
			
			E17 & 5 & 4 & 1 & 3 & 100 000 \\
			
			E18 & 6 & 0 & 6 & 1 & 50 000 \\
			
			E19 & 2 & 0 & 2 & 10 & 50 000 \\
			
			E20 & 2 & 0 & 2 & 9 & 50 000 \\
			
			E21 & 2 & 0 & 2 & 13 & 50 000 \\
			
			E22 & 8 & 0 & 8 & 16 & 100 000 \\
			
			E23 & 2 & 0 & 2 & 6 & 50 000 \\
			
			E24 & 20 & 19 & 1 & 2 & 200 000 \\
			
			E25 & 3 & 0 & 3 & 7 & 50 000 \\
			
			E26 & 2 & 0 & 2 & 4 & 50 000 \\
			
			E27 & 2 & 0 & 2 & 6 & 50 000 \\
			
			E28 & 3 & 0 & 3 & 8 & 100 000 \\
			
			E29 & 3 & 0 & 3 & 2 & 50 000 \\
			
			E30 & 3 & 0 & 3 & 12 & 50 000 \\
			
			E31 & 2 & 0 & 2 & 2 & 50 000 \\
			
			E32 & 2 & 0 & 2 & 4 & 50 000 \\
			
			E33 & 2 & 0 & 2 & 4 & 50 000 \\
			
			E34 & 2 & 0 & 2 & 2 & 50 000 \\
			
			E35 & 3 & 0 & 3 & 1 & 50 000 \\
			
			E36 & 2 & 0 & 2 & 2 & 50 000 \\
			
			E37 & 3 & 0 & 3 & 1 & 50 000 \\
			
			E38 & 2 & 0 & 2 & 3 & 50 000 \\
			
			E39 & 2 & 0 & 2 & 2 & 50 000 \\
			
			E40 & 3 & 0 & 3 & 5 & 50 000 \\
			
			E41 & 2 & 0 & 2 & 4 & 50 000 \\
			
			E42 & 2 & 0 & 2 & 2 & 50 000 \\
			
			E43 & 5 & 0 & 5 & infinite & 200 000 \\
			
			E44 & 6 & 0 & 6 & infinite & 200 000 \\
			
			E45 & 10 & 4 & 6 & infinite & 200 000 \\
			
			E46 & 5 & 1 & 4 & infinite & 200 000 \\
			
			\bottomrule

		\end{tabular}
	\end{center}
\end{table}

\subsubsection{Performance metrics}

In this section, two performance metrics inspired by the multi-model and multi-objective problems are adopted to comprehensively assess the solutions found by a method. Besides, one performance metric is employed to evaluate the quality of the roots found by a method.

\begin{enumerate}
	\item Root ratio (RR) \cite{thomsen2004multimodal}: RR computes the average ratio of roots found over multiple runs:
	\begin{equation}\label{Eq:RR}
	{\rm{RR}} = \frac{{\sum\nolimits_{i = 1}^{{N_r}} {{N_{f,i}}} }}{{NoR \cdot {N_r}}},
	\end{equation}
	where ${N_r}$ is the number of runs. ${N_{f,i}}$ is the number of roots found in the $i$-th run and $NoR$ is the number of actual known roots of a NES. In this section, for a solution $\vec x$, if its repulsion value $R(\vec x) < 1e - 5$, it can be regarded as a root \cite{liao2018solving}. To make the experimental results general and reliable, each algorithm is executed over ${N_r}{\rm{ = 3}}0$ independent runs for each NES.

	\item Success rate (SR) \cite{li2013benchmark}: SR measures the ratio of successful runs. A successful run refers to a run where all the actual known roots of a NES are found, the expression of SR is:
	\begin{equation}\label{Eq:SR}
	{\rm{SR}} = \frac{{{N_{r,s}}}}{{{N_r}}},
	\end{equation}
	where ${N_{r,s}}$ is the number of successful runs.

	The optimal solutions of NESs E43-E46 are unknown, so the RR and SR criteria cannot be used for the performance evaluation. We will discuss the performance of NESs E43-E46 in the next section.

	\item Evaluating the quality of roots found (QR): The mean of the objective function values $\sum\nolimits_{j = 1}^m {f_j^2(\vec x)} $ for the $i$-th run:
	\begin{equation}\label{Eq:QR}
	{\rm{QR = }}\sum\nolimits_{l = 1}^{{N_{f,i}}} {\sum\nolimits_{j = 1}^m {f_j^2({{\vec x}_l})} } /{N_{f,i}},
	\end{equation}
	
	Where $m$ is the number of equations for a NES, ${\vec x_l}$ is the $l$-th of roots found in the $i$-th run. QR indicator adapted from IGD indicator, which can measure the quality of the roots obtained by an algorithm.
\end{enumerate}

\subsubsection{Variable reduction results}

Due to space limitation, the expressions of the 46 NESs and the selected reduction schemes are shown in the supplementary file.

Among the reduction schemes of the 46 NESs, except E5, E12, and E21, all the NESs can be reduced by VRS. A NES may have more than one reduction scheme. We show one promising reduction scheme for each NES in the supplementary file. For the NESs that cannot be reduced, they can be divided into two categories:

\begin{enumerate}
	\item No variable in a NES can be explicitly expressed by other variables, such as E5 and E21.
	
	\item There is a periodic function in a NES, and no variable can be completely represented and calculated by other variables in the NES, such as E12.
\end{enumerate}

\subsubsection{Experimental results and discussion on E1-E46}

Except NESs E5, E12, E21 that cannot be reduced, the experimental results of DR-JADE and VR-DR-JADE on the other 39 NESs concerning RR-indicator, SR-indicator and QR-indicator over 30 independent runs are reported in Table~\ref{Table:results of VR-DR-JADE}.
{
	\aboverulesep=0pt \belowrulesep=0pt
	\begin{table*}[htp]
	\renewcommand{\arraystretch}{0.95}
	\caption{\label{Table:results of VR-DR-JADE}Status of QR-indicator, RR-indicator and SR-indicator of DR-JADE and VR-DR-JADE, where “NaN” means not available.}
	\footnotesize
	
	\begin{center}
		\begin{tabular}{cccccccc}\toprule

			\multirow{2}{*}{Test problem} & \multirow{2}{*}{Status} & \multicolumn{2}{c}{QR} & \multicolumn{2}{c}{RR} & \multicolumn{2}{c}{SR} \\
			\cline{3-8}
			&   & DR-JADE & VR-DR-JADE & DR-JADE & VR-DR-JADE & DR-JADE & VR-DR-JADE \\
			\midrule		  
			
			\multirow{2}{*}{E1} & Mean & 2.35E-06 & {\bf 4.12E-07} & \multirow{2}{*}{1.0000} & \multirow{2}{*}{\bf 1.0000}  & \multirow{2}{*}{1.0000}  & \multirow{2}{*}{\bf 1.0000} \\
			& Std & 1.52E-06 & {\bf 8.41E-07}  \\	 \hline

			\multirow{2}{*}{E2} & Mean & 5.24E-16 & {\bf 9.56E-22} & \multirow{2}{*}{1.0000} & \multirow{2}{*}{\bf 1.0000}  & \multirow{2}{*}{1.0000}  & \multirow{2}{*}{\bf 1.0000} \\
			& Std & 2.15E-15 & {\bf 5.06E-21}  \\	 \hline

			\multirow{2}{*}{E3} & Mean & 3.12E-15 & {\bf 1.36E-16} & \multirow{2}{*}{0.8667} & \multirow{2}{*}{\bf 1.0000}  & \multirow{2}{*}{0.7333}  & \multirow{2}{*}{\bf 1.0000} \\
			& Std & 1.11E-14 & {\bf 5.18E-16}  \\	 \hline

			\multirow{2}{*}{E4} & Mean & 1.05E-07 & {\bf 2.10E-31} & \multirow{2}{*}{1.0000} & \multirow{2}{*}{\bf 1.0000}  & \multirow{2}{*}{1.0000}  & \multirow{2}{*}{\bf 1.0000} \\
			& Std & 5.74E-07 & {\bf 1.15E-30}  \\	 \hline

			\multirow{2}{*}{E6} & Mean & 8.72E-24 & {\bf 0.00E+00} & \multirow{2}{*}{1.0000} & \multirow{2}{*}{\bf 1.0000}  & \multirow{2}{*}{1.0000}  & \multirow{2}{*}{\bf 1.0000} \\
			& Std & 4.02E-23 & {\bf 0.00E+00}  \\	 \hline

			\multirow{2}{*}{E7} & Mean & 7.79E-20 & {\bf 6.70E-31} & \multirow{2}{*}{1.0000} & \multirow{2}{*}{\bf 1.0000}  & \multirow{2}{*}{1.0000}  & \multirow{2}{*}{\bf 1.0000} \\
			& Std & 4.24E-19 & {\bf 3.36E-30}  \\	 \hline

			\multirow{2}{*}{E8} & Mean & 2.39E-20 & {\bf 1.74E-24} & \multirow{2}{*}{1.0000} & \multirow{2}{*}{\bf 1.0000}  & \multirow{2}{*}{1.0000}  & \multirow{2}{*}{\bf 1.0000} \\
			& Std & 1.13E-19 & {\bf 9.52E-24}  \\	 \hline

			\multirow{2}{*}{E9} & Mean & NaN & {\bf NaN} & \multirow{2}{*}{0.0000} & \multirow{2}{*}{\bf 0.0000}  & \multirow{2}{*}{0.0000}  & \multirow{2}{*}{\bf 0.0000} \\
			& Std & NaN & {\bf NaN}  \\	 \hline

			\multirow{2}{*}{E10} & Mean & 3.76E-08 & {\bf 7.68E-11} & \multirow{2}{*}{0.9970} & \multirow{2}{*}{\bf 1.0000}  & \multirow{2}{*}{0.9667}  & \multirow{2}{*}{\bf 1.0000} \\
			& Std & 1.17E-07 & {\bf 4.08E-10}  \\	 \hline

			\multirow{2}{*}{E11} & Mean & 7.02E-09 & {\bf 1.28E-09} & \multirow{2}{*}{0.9378} & \multirow{2}{*}{\bf 1.0000}  & \multirow{2}{*}{0.2667}  & \multirow{2}{*}{\bf 1.0000} \\
			& Std & 3.24E-08 & {\bf 6.97E-09}  \\	 \hline

			\multirow{2}{*}{E13} & Mean & 1.93E-06 & {\bf 8.30E-12} & \multirow{2}{*}{1.0000} & \multirow{2}{*}{\bf 1.0000}  & \multirow{2}{*}{1.0000}  & \multirow{2}{*}{\bf 1.0000} \\
			& Std & 3.24E-08 & {\bf 6.97E-09}  \\	 \hline

			\multirow{2}{*}{E14} & Mean & 4.05E-09 & {\bf 9.32E-16} & \multirow{2}{*}{0.9667} & \multirow{2}{*}{\bf 1.0000}  & \multirow{2}{*}{0.7333}  & \multirow{2}{*}{\bf 1.0000} \\
			& Std & 1.61E-08 & {\bf 4.07E-15}  \\	 \hline

			\multirow{2}{*}{E15} & Mean & 8.28E-08 & {\bf 1.03E-28} & \multirow{2}{*}{1.0000} & \multirow{2}{*}{\bf 1.0000}  & \multirow{2}{*}{1.0000}  & \multirow{2}{*}{\bf 1.0000} \\
			& Std & 4.53E-07 & {\bf 8.72E-29}  \\	 \hline

			\multirow{2}{*}{E16} & Mean & 8.02E-08 & {\bf 4.69E-09} & \multirow{2}{*}{1.0000} & \multirow{2}{*}{0.8571}  & \multirow{2}{*}{1.0000}  & \multirow{2}{*}{0.0000} \\
			& Std & 2.60E-07 & {\bf 1.62E-08}  \\	 \hline

			\multirow{2}{*}{E17} & Mean & 3.28E-13 & {\bf 8.05E-31} & \multirow{2}{*}{1.0000} & \multirow{2}{*}{\bf 1.0000}  & \multirow{2}{*}{1.0000}  & \multirow{2}{*}{\bf 1.0000} \\
			& Std & 1.19E-12 & {\bf 3.56E-46}  \\	 \hline

			\multirow{2}{*}{E18} & Mean & 9.78E-32 & {\bf 0.00E+00} & \multirow{2}{*}{1.0000} & \multirow{2}{*}{\bf 1.0000}  & \multirow{2}{*}{1.0000}  & \multirow{2}{*}{\bf 1.0000} \\
			& Std & 3.16E-31 & {\bf 0.00E+00}  \\	 \hline

			\multirow{2}{*}{E19} & Mean & 4.61E-09 & {\bf 1.70E-20} & \multirow{2}{*}{0.8600} & \multirow{2}{*}{\bf 1.0000}  & \multirow{2}{*}{0.0333}  & \multirow{2}{*}{\bf 1.0000} \\
			& Std & 1.56E-08 & {\bf 9.31E-20}  \\	 \hline

			\multirow{2}{*}{E20} & Mean & 3.43E-09 & {\bf 3.30E-13} & \multirow{2}{*}{1.0000} & \multirow{2}{*}{\bf 1.0000}  & \multirow{2}{*}{1.0000}  & \multirow{2}{*}{\bf 1.0000} \\
			& Std & 1.83E-08 & {\bf 1.15E-12}  \\	 \hline

			\multirow{2}{*}{E22} & Mean & 7.67E-11 & {\bf 6.09E-27} & \multirow{2}{*}{0.8375} & \multirow{2}{*}{\bf 0.8729}  & \multirow{2}{*}{0.0333}  & \multirow{2}{*}{\bf 0.0667} \\
			& Std & 2.90E-10 & {\bf 2.69E-26}  \\	 \hline

			\multirow{2}{*}{E23} & Mean & 3.51E-13 & {\bf 7.11E-29} & \multirow{2}{*}{1.0000} & \multirow{2}{*}{\bf 1.0000}  & \multirow{2}{*}{1.0000}  & \multirow{2}{*}{\bf 1.0000} \\
			& Std & 1.92E-12 & {\bf 3.83E-28}  \\	 \hline

			\multirow{2}{*}{E24} & Mean & NaN & {\bf 0.00E+00} & \multirow{2}{*}{0.0000} & \multirow{2}{*}{\bf 1.0000}  & \multirow{2}{*}{0.0000}  & \multirow{2}{*}{\bf 1.0000} \\
			& Std & NaN & {\bf 0.00E+00}  \\	 \hline

			\multirow{2}{*}{E25} & Mean & 3.16E-16 & 2.04E-14 & \multirow{2}{*}{1.0000} & \multirow{2}{*}{\bf 1.0000}  & \multirow{2}{*}{1.0000}  & \multirow{2}{*}{\bf 1.0000} \\
			& Std & 1.27E-15 & 1.12E-13  \\	 \hline

			\multirow{2}{*}{E26} & Mean & 3.46E-12 & {\bf 4.74E-27} & \multirow{2}{*}{1.0000} & \multirow{2}{*}{\bf 1.0000}  & \multirow{2}{*}{1.0000}  & \multirow{2}{*}{\bf 1.0000} \\
			& Std & 1.89E-11 & {\bf 2.60E-26}  \\	 \hline

			\multirow{2}{*}{E27} & Mean & 3.15E-11 & {\bf 9.58E-18} & \multirow{2}{*}{1.0000} & \multirow{2}{*}{\bf 1.0000}  & \multirow{2}{*}{1.0000}  & \multirow{2}{*}{\bf 1.0000} \\
			& Std & 1.72E-10 & {\bf 5.23E-17}  \\	 \hline

			\multirow{2}{*}{E28} & Mean & 2.39E-07 & {\bf 1.51E-32} & \multirow{2}{*}{0.9208} & \multirow{2}{*}{\bf 1.0000}  & \multirow{2}{*}{0.5000}  & \multirow{2}{*}{\bf 1.0000} \\
			& Std & 1.90E-07 & {\bf 1.54E-33}  \\	 \hline

			\multirow{2}{*}{E29} & Mean & 4.93E-15 & {\bf 1.55E-18} & \multirow{2}{*}{1.0000} & \multirow{2}{*}{\bf 1.0000}  & \multirow{2}{*}{1.0000}  & \multirow{2}{*}{\bf 1.0000} \\
			& Std & 2.69E-14 & {\bf 8.45E-18}  \\	 \hline

			\multirow{2}{*}{E30} & Mean & 6.35E-09 & {\bf 3.92E-09} & \multirow{2}{*}{0.9306} & \multirow{2}{*}{0.8833}  & \multirow{2}{*}{0.3667}  & \multirow{2}{*}{0.3333} \\
			& Std & 1.30E-08 & {\bf 9.38E-09}  \\	 \hline

			\multirow{2}{*}{E31} & Mean & 1.61E-21 & {\bf 2.99E-48} & \multirow{2}{*}{1.0000} & \multirow{2}{*}{\bf 1.0000}  & \multirow{2}{*}{1.0000}  & \multirow{2}{*}{\bf 1.0000} \\
			& Std & 8.82E-21 & {\bf 1.64E-47}  \\	 \hline

			\multirow{2}{*}{E32} & Mean & 1.54E-14 & {\bf 4.45E-20} & \multirow{2}{*}{1.0000} & \multirow{2}{*}{\bf 1.0000}  & \multirow{2}{*}{1.0000}  & \multirow{2}{*}{\bf 1.0000} \\
			& Std & 8.41E-14 & {\bf 2.43E-19}  \\	 \hline

			\multirow{2}{*}{E33} & Mean & 1.11E-15 & {\bf 7.39E-19} & \multirow{2}{*}{1.0000} & \multirow{2}{*}{\bf 1.0000}  & \multirow{2}{*}{1.0000}  & \multirow{2}{*}{\bf 1.0000} \\
			& Std & 4.66E-15 & {\bf 4.05E-18}  \\	 \hline

			\multirow{2}{*}{E34} & Mean & 4.98E-07 & {\bf 8.73E-26} & \multirow{2}{*}{0.5000} & \multirow{2}{*}{\bf 1.0000}  & \multirow{2}{*}{0.0000}  & \multirow{2}{*}{\bf 1.0000} \\
			& Std & 7.16E-07 & {\bf 4.78E-25}  \\	 \hline

			\multirow{2}{*}{E35} & Mean & 2.16E-17 & {\bf 3.02E-18} & \multirow{2}{*}{1.0000} & \multirow{2}{*}{\bf 1.0000}  & \multirow{2}{*}{1.0000}  & \multirow{2}{*}{\bf 1.0000} \\
			& Std & 6.56E-17 & {\bf 1.13E-17}  \\	 \hline

			\multirow{2}{*}{E36} & Mean & 1.89E-19 & {\bf 4.72E-29} & \multirow{2}{*}{1.0000} & \multirow{2}{*}{\bf 1.0000}  & \multirow{2}{*}{1.0000}  & \multirow{2}{*}{\bf 1.0000} \\
			& Std & 1.03E-18 & {\bf 2.58E-28}  \\	 \hline

			\multirow{2}{*}{E37} & Mean & 9.56E-09 & {\bf 2.31E-09} & \multirow{2}{*}{1.0000} & \multirow{2}{*}{\bf 1.0000}  & \multirow{2}{*}{1.0000}  & \multirow{2}{*}{\bf 1.0000} \\
			& Std & 4.91E-08 & {\bf 1.26E-08}  \\	 \hline

			\multirow{2}{*}{E38} & Mean & 3.05E-19 & {\bf 9.57E-24} & \multirow{2}{*}{1.0000} & \multirow{2}{*}{\bf 1.0000}  & \multirow{2}{*}{1.0000}  & \multirow{2}{*}{\bf 1.0000} \\
			& Std & 1.08E-18 & {\bf 4.81E-23}  \\	 \hline

			\multirow{2}{*}{E39} & Mean & 8.40E-09 & {\bf 2.33E-25} & \multirow{2}{*}{1.0000} & \multirow{2}{*}{\bf 1.0000}  & \multirow{2}{*}{1.0000}  & \multirow{2}{*}{\bf 1.0000} \\
			& Std & 4.60E-08 & {\bf 0.00E+00}  \\	 \hline

			\multirow{2}{*}{E40} & Mean & 8.82E-09 & {\bf 3.05E-09} & \multirow{2}{*}{0.9467} & \multirow{2}{*}{\bf 0.9867}  & \multirow{2}{*}{0.7333}  & \multirow{2}{*}{\bf 0.9333} \\
			& Std & 4.81E-08 & {\bf 1.67E-08}  \\	 \hline

			\multirow{2}{*}{E41} & Mean & 3.73E-20 & {\bf 3.94E-31} & \multirow{2}{*}{1.0000} & \multirow{2}{*}{\bf 1.0000}  & \multirow{2}{*}{1.0000}  & \multirow{2}{*}{\bf 1.0000} \\
			& Std & 1.81E-19 & {\bf 0.00E+00}  \\	 \hline

			\multirow{2}{*}{E42} & Mean & 1.23E-21 & {\bf 1.83E-28} & \multirow{2}{*}{1.0000} & \multirow{2}{*}{\bf 1.0000}  & \multirow{2}{*}{1.0000}  & \multirow{2}{*}{\bf 1.0000} \\
			& Std & 6.73E-21 & {\bf 7.00E-28}  \\	 
			\bottomrule
			
		\end{tabular}
	\end{center}
\end{table*}
}

The major purpose of DR-JADE is to obtain multiple roots of a NES, while the quality of the roots is slightly neglected. From the results in Table~\ref{Table:results of VR-DR-JADE}, compared VR-DR-JADE with DR-JADE, except a slight increase for E25, the mean and standard deviation values of QR-indicator values for each test problem in E1-E42 (except E5, E12 and E21) have decreased. The above phenomenon indicates that the most roots located by VR-DR-JADE are closer to the actual known roots than the roots located by DR-JADE. Additionally, to further compare the quality of the overall solutions of VR-DR-JADE and DR-JADE, the Wilcoxon test on the mean QR-indicator in Table~\ref{Table:results of VR-DR-JADE} is performed. Compared VR-DR-JADE with DR-JADE, we can get ${R^ + } = 687.0$, ${R^ - } = 16.0$ and $p = 2.46{\rm{E - 09}}$. Since VR-DR-JADE can provide higher ${R^ + }$ value than ${R^ - }$ value and the $p$ value is less than 0.05, VR-DR-JADE significantly outperforms DR-JADE in terms of the overall quality of solutions.

VR-DR-JADE can get 10 better and 27 equal values in both RR and SR compared with DR-JADE for NESs E1-E42 (except E5, E12 and E21). It is worth noting that VR-DR-JADE can locate all the roots of NES E24 over each run. In contrast, DR-JADE cannot locate any roots for NES E24 over 30 runs. For NES E9, VR-DR-JADE and DR-JADE both cannot find any roots. But when the number of the decision variables of NES E9 is set to 10, VR-DR-JADE can find all the roots of NES E9 while DR-JADE still cannot find any roots.

For NES E16, VR-DR-JADE always has one root not found over each run, which may be related to the search ability of DR-JADE itself. The RR and SR values of VR-DR-JADE is slightly lower than DR-JADE’s for NES E30. The reason may be that the adoption of VRS makes the great majority of reduced variable values violate the reduced variable's boundary constraint during the evolution, which greatly reduces the number of feasible solutions.

To further study the roots obtained by VR-DR-JADE, we compare VR-DR-JADE with ten peer methods, i.e., DR-JADE \cite{liao2018solving}, A-WeB[33] \cite{gong2017weighted}, VR-MONES, MONES \cite{song2014locating}, I-HS \cite{ramadas2014multiple}, NCDE \cite{qu2012differential}, NSDE \cite{qu2012differential}, GA-SQP \cite{raja2016memetic}, PSO-NM \cite{raja2018nature}, and NCSA \cite{zhang2019applying}\footnote{Except for VR-DR-JADE, DR-JADE, VR-MONES, and MONES, the data of other seven compared methods are from the literature \cite{liao2018solving} and the detailed results of the eleven methods are reported in the supplementary file.}. Note that if we map the individuals in a population to the objective space defined by MONES and VR-MONES for NESs E1-E42, the roots that have the same values in the first decision variable will be considered to be discovered even if only a few of them are located. To address this issue, we propose another way to judge whether an individual in the final population is a root or not, i.e., when the minimum Euclidean distance between an individual in the final population and the individuals in the set of optimal solutions for a NES is less than 0.01, the individual can be considered as a root of the NES. Table~\ref{Table:Friedman test} shows the Friedman test of RR and SR for the 11 methods. Table~\ref{Table:Wilcoxon test} displays the Wilcoxon test obtained by the comparison of VR-DR-JADE and the other ten methods for both RR and SR.
\aboverulesep=0pt \belowrulesep=0pt
\begin{table}[htp]
	\footnotesize
	\caption{Average rankings of VR-DR-JADE and the other ten state-of-the-art algorithms obtained by the Friedman test for both RR and SR}
	\label{Table:Friedman test}
	\begin{center}
		\begin{tabular}{ccc} \toprule
			Algorithm & Ranking (RR) & Ranking (SR) \\ \midrule	
			
			VR-DR-JADE  &{\bf 3.2308} & {\bf 3.2564} \\
			DR-JADE & 3.8718 & 3.9103 \\
			VR-MONES & 4.6538 & 4.8077 \\
			MONES & 7.1282 & 7.1795 \\ 
			A-WeB & 5.1282 & 5.0641 \\
			NCDE & 5.7821 & 5.6410 \\ 
			NSDE & 5.2564 & 5.3333 \\ 
			I-HS & 6.1667 & 6.2949 \\ 
			GA-SQP & 9.8974 & 9.4359 \\ 
			PSO-NM & 6.8974  & 7.0256\\
			NCSA & 7.9872 & 8.0513 \\
			
			\bottomrule
			
		\end{tabular}
	\end{center}
\end{table}
\aboverulesep=0pt \belowrulesep=0pt
\begin{table}[htp]
	\footnotesize
	\renewcommand{\arraystretch}{1.1}
	\centering
	\caption{Results of VR-DR-JADE compared with the other ten state-of-the-art algorithms obtained by the Wilcoxon test for both RR and SR}
	\label{Table:Wilcoxon test}
	\begin{center}
		\begin{tabular}{ccccccc} \toprule
			\multirow{2}{*}{\tabincell{c}{VR-DR-JADE \\ $VS$}} & \multicolumn{3}{c}{RR} & \multicolumn{3}{c}{SR}  \\
			\cline{2-7}
			& ${R^ + }$ & ${R^ - }$ & $p$-value & ${R^ + }$ & ${R^ - }$ & $p$-value  \\
			\midrule

			DR-JADE & 498.5 & 242.5 & 6.40E-02 & 528.5 & 212.5 & {\bf 2.11E-02}  \\
			
			VR-MONE & 528.5 & 212.5 & {\bf 2.11E-02} & 529.0 & 212.0 & {\bf 2.07E-02}  \\
			
			MONES & 665.0 & 76.0 & {\bf 3.76E-06} & 696.0 & 84.0 & {\bf 4.02E-06}  \\
			
			A-WeB & 674.0 & 106.0 & {\bf 2.57E-05} & 660.0 & 120.0 & {\bf 7.35E-05} \\
			
			NCDE & 597.0 & 144.0 & {\bf 6.83E-04} & 593.0 & 148.0 & {\bf 8.70E-04}  \\
			
			NSDE & 625.5 & 154.5 & {\bf 6.87E-04} & 623.5 & 156.5 & {\bf 7.72E-04}  \\
			
			I-HS & 670.0 & 71.0 & {\bf 2.30E-06} & 710.5 & 69.5 & {\bf 9.98E-07}  \\
			
			GA-SQP & 778.5 & 1.5 & {\bf 9.09E-12} & 739.5 & 1.5 & {\bf 1.82E-11}  \\
			
			PSO-NM & 701.5 & 39.5 & {\bf 5.92E-08} & 695.0 & 46.0 & {\bf 1.39E-07}  \\
			
			NCSA & 735.0 & 45.0 & {\bf 6.12E-08} & 717.5 & 23.5 & {\bf 5.10E-09}   \\         
			
			\bottomrule

		\end{tabular}
	\end{center}
\end{table}

From Table~\ref{Table:Friedman test}, VR-DR-JADE has achieved the highest Friedman test rankings for both SR and RR. What's more, VR-MONES can also obtain higher Friedman test rankings for both SR and RR than MONES. The above results reveal that the integration of VRS can make the algorithms locate more roots than the original algorithms overall. Meanwhile, Table~\ref{Table:Wilcoxon test} shows that VR-DR-JADE significantly outperforms the other ten methods for RR and SR by the Wilcoxon test, since it can provide higher ${R^ + }$ values than ${R^ - }$ values in all cases and all $p$ values are less than 0.05 except the $p$ values obtained by VR-DR-JADE and DR-JADE. Therefore, we can conclude that the integration of VRS can be an effective way to improve the performance of the existing method.

\begin{figure*}[htp]
	\begin{center}
		\subfigure[\small{t=1}]{\psfig{file=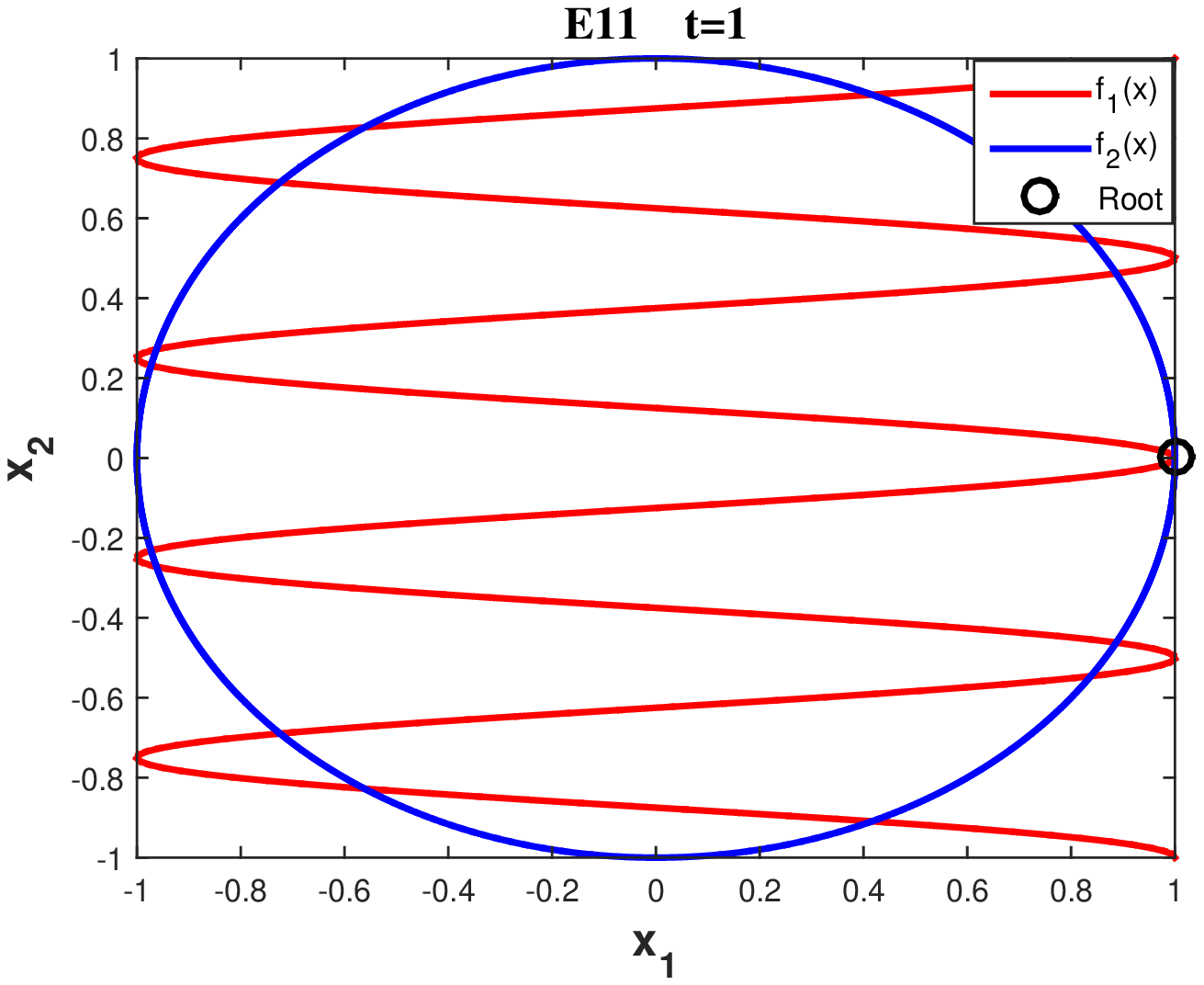,width=1.55in}}
		\subfigure[\small{t=5}]{\psfig{file=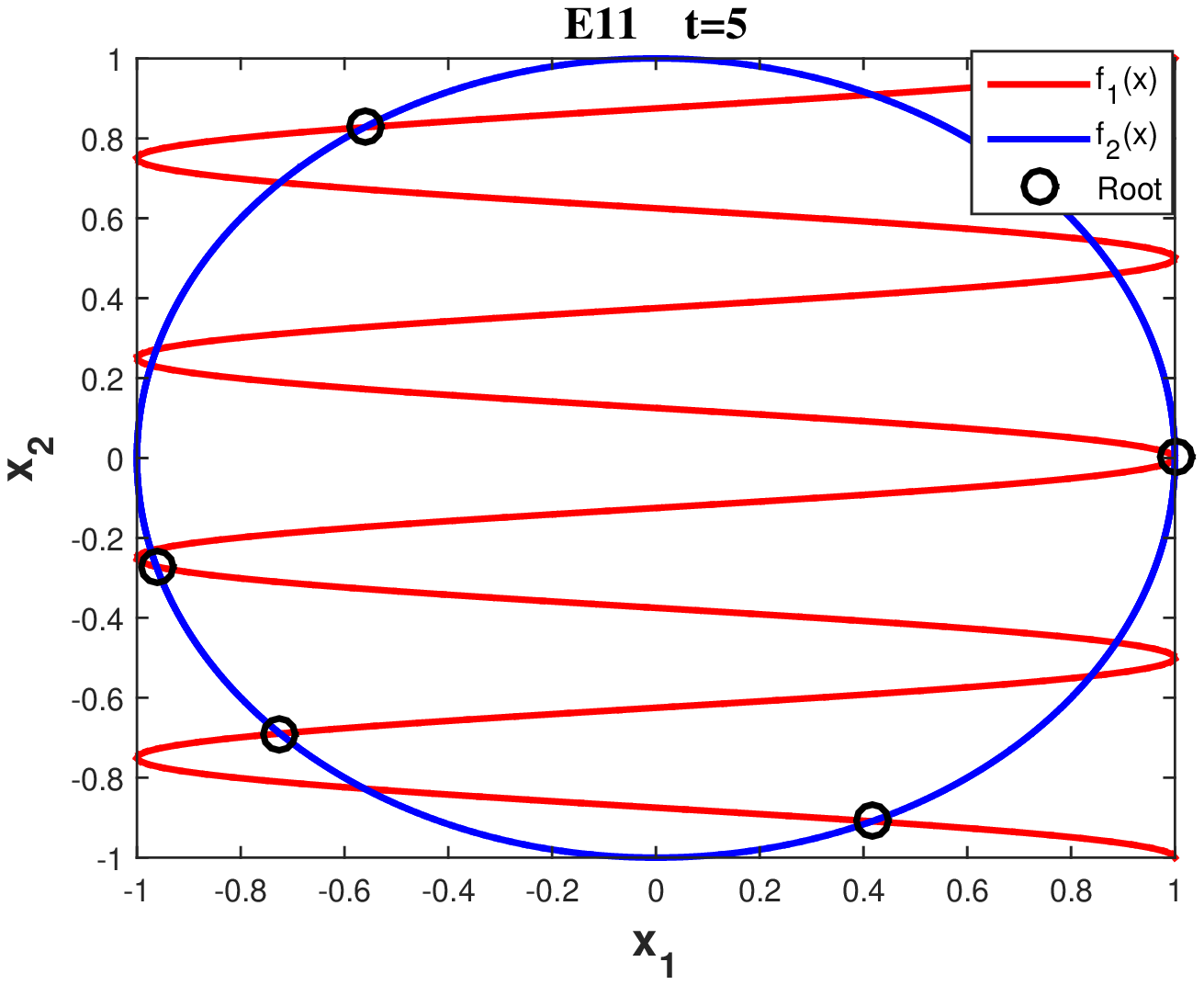,width=1.55in}}
		\subfigure[\small{t=15}]{\psfig{file=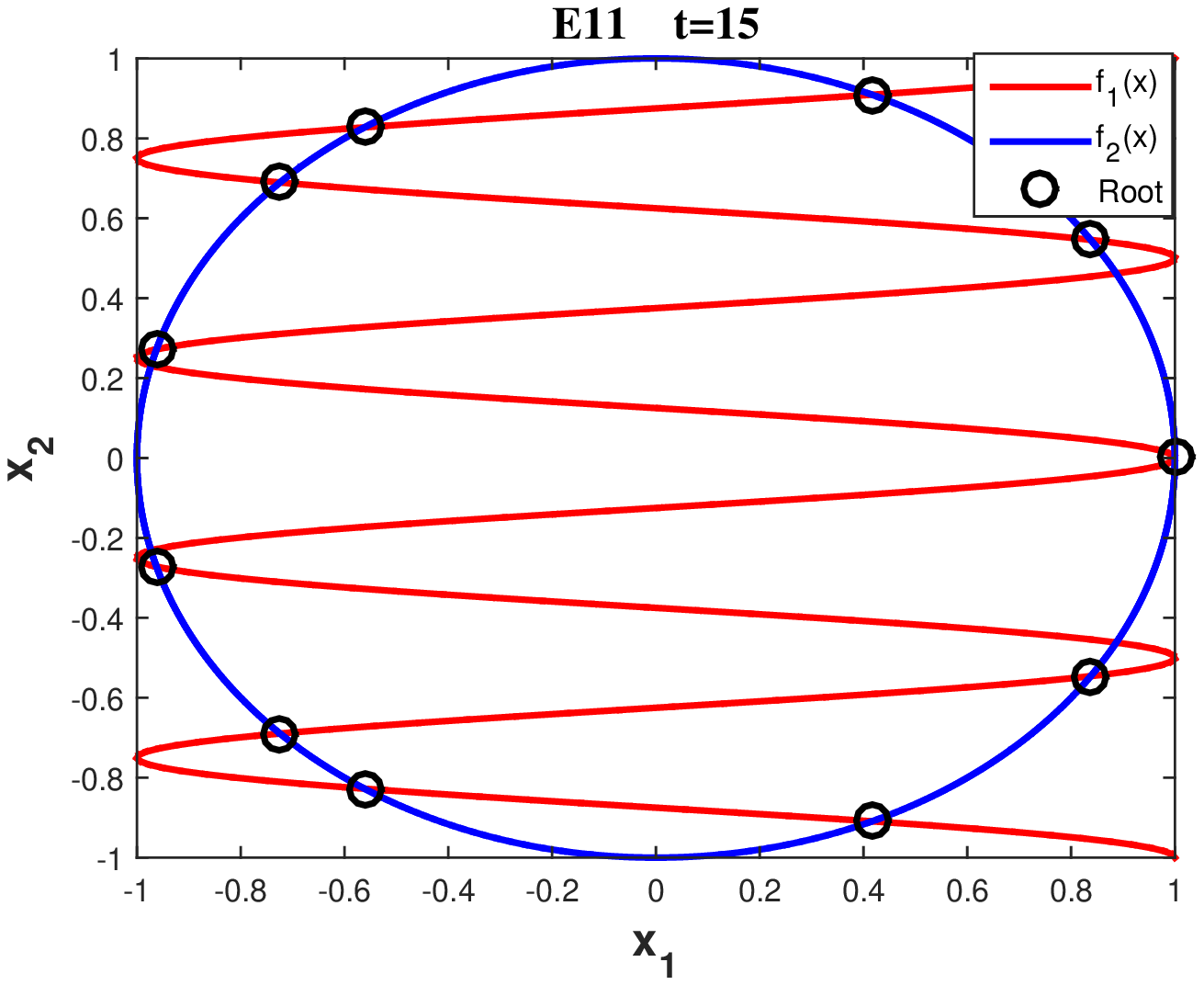,width=1.55in}}
		\subfigure[\small{t=30}]{\psfig{file=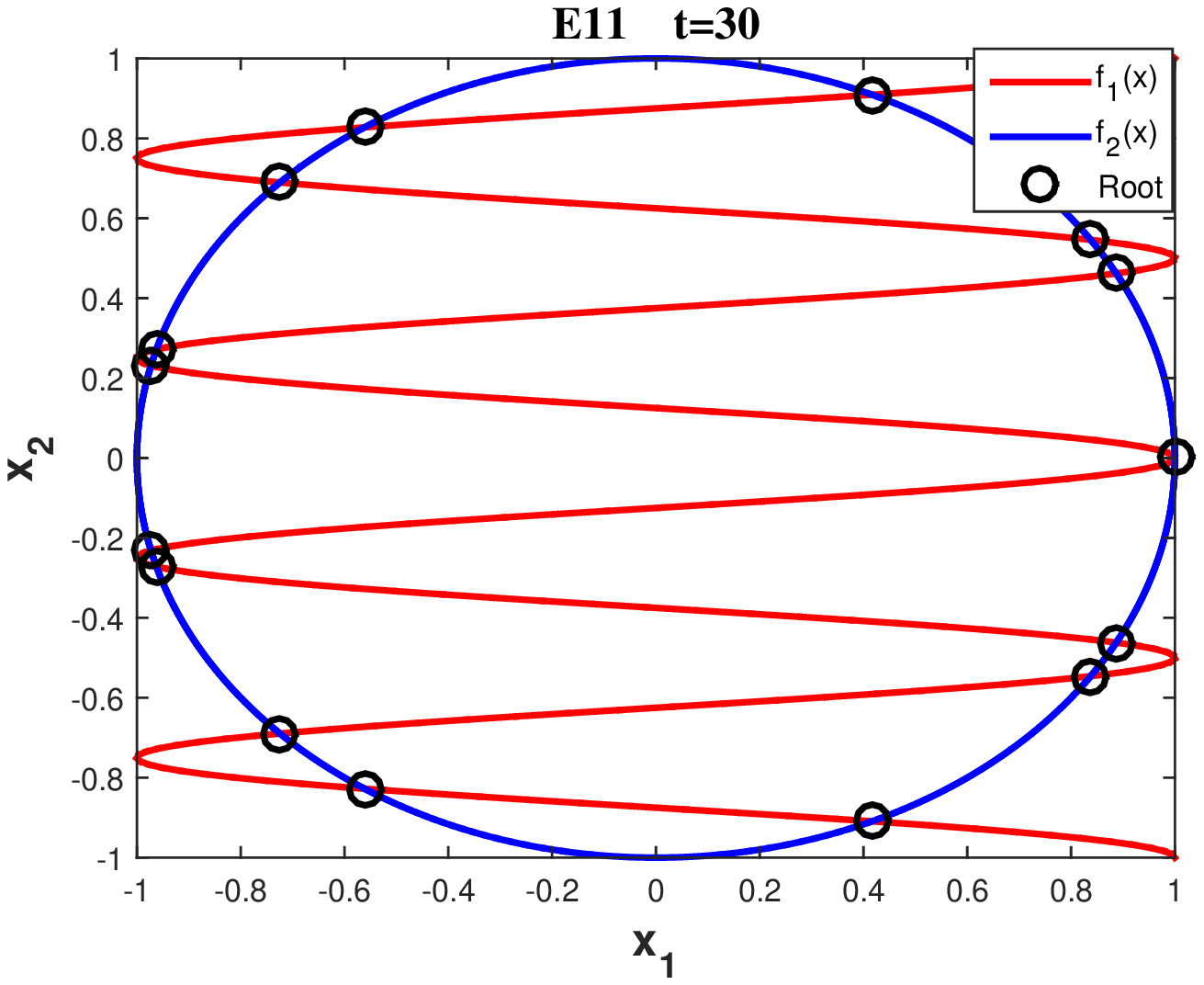,width=1.55in}}
	\end{center}\vspace{-3mm}
	\caption{Evolution of VR-DR-JADE over a typical run on E11} \label{Fig:Evolution of VE11}
\end{figure*}

\begin{figure*}[htp]
	\begin{center}
		\subfigure[\small{t=1}]{\psfig{file=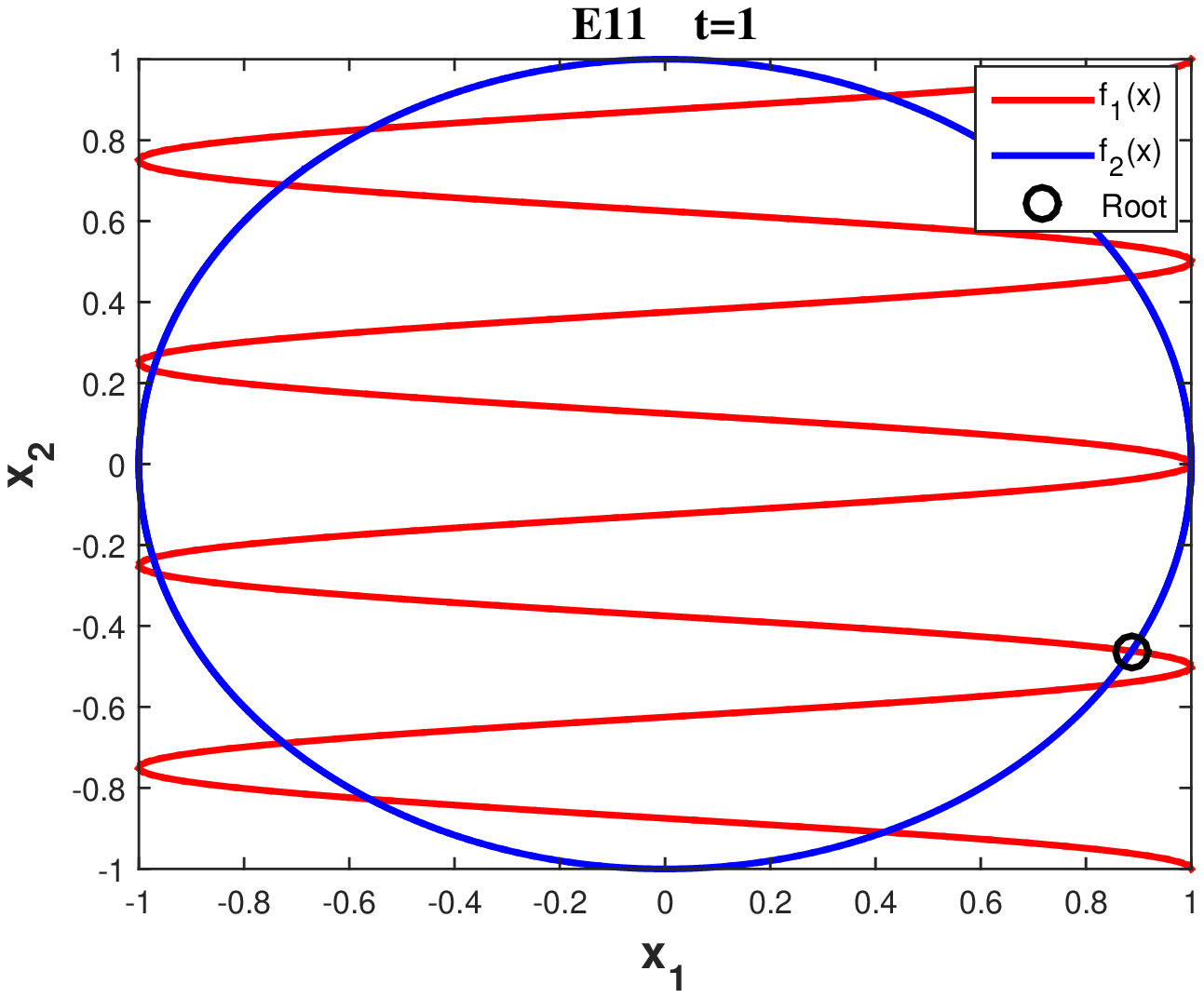,width=1.55in}}
		\subfigure[\small{t=5}]{\psfig{file=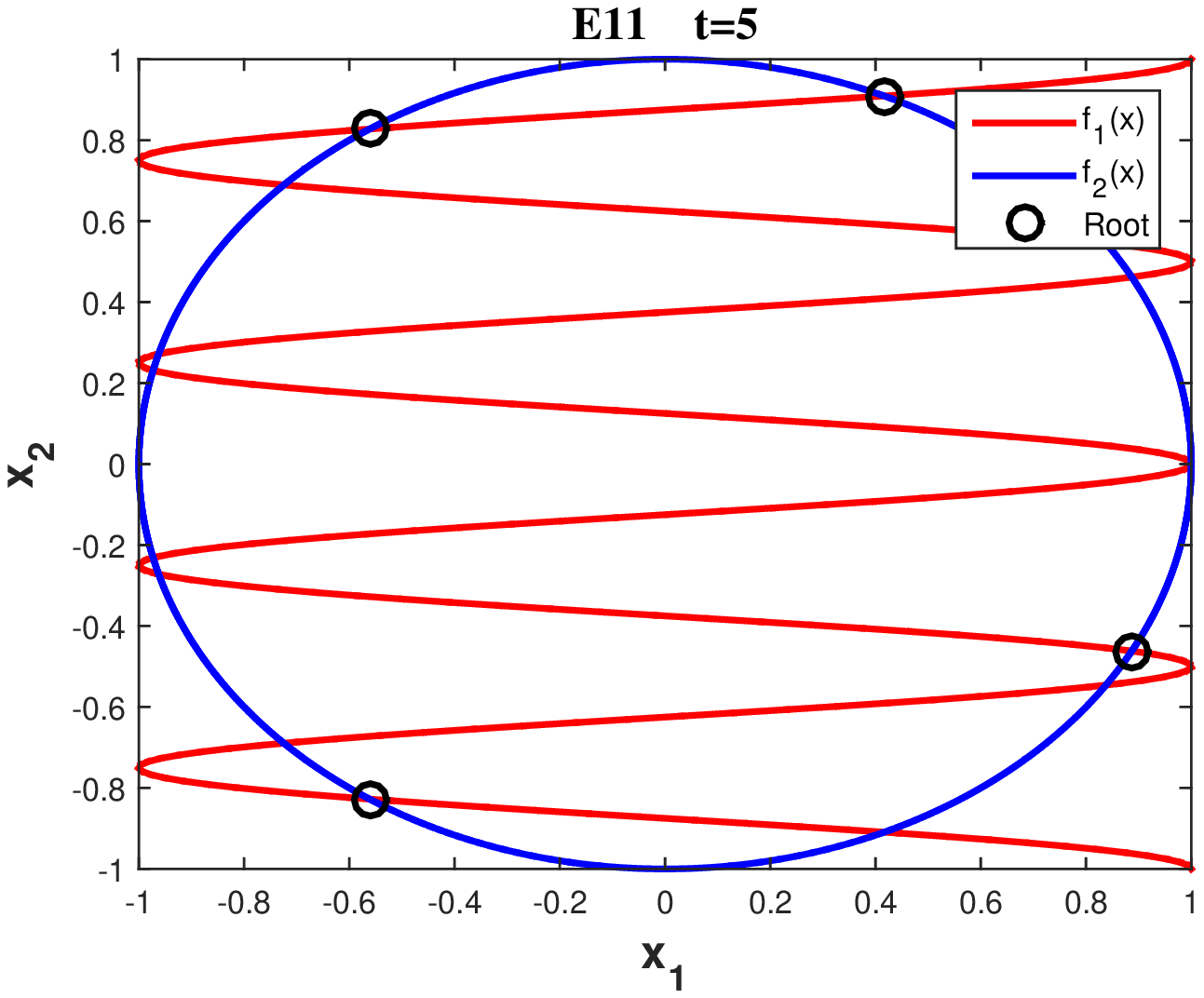,width=1.55in}}
		\subfigure[\small{t=15}]{\psfig{file=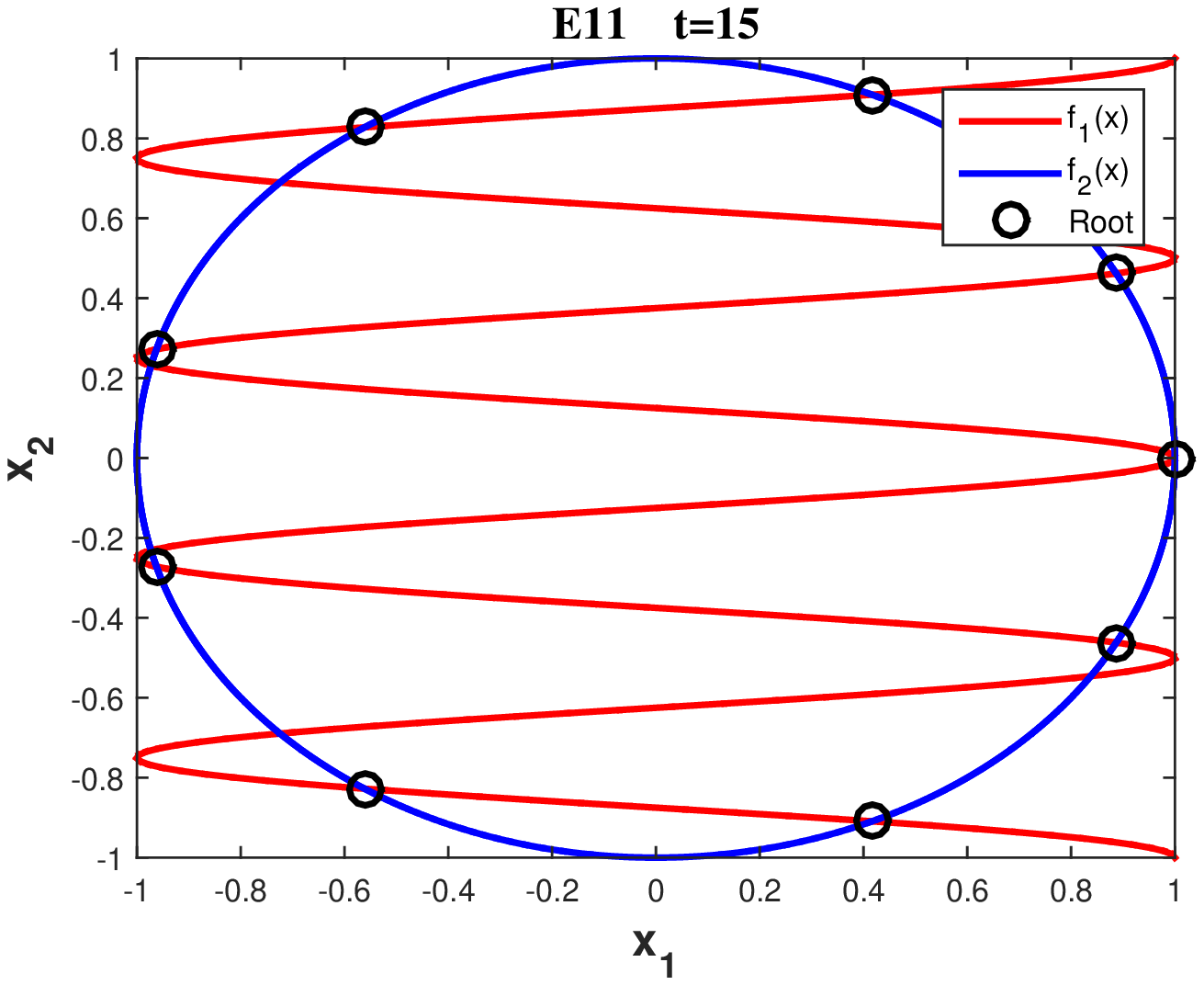,width=1.55in}}
		\subfigure[\small{t=30}]{\psfig{file=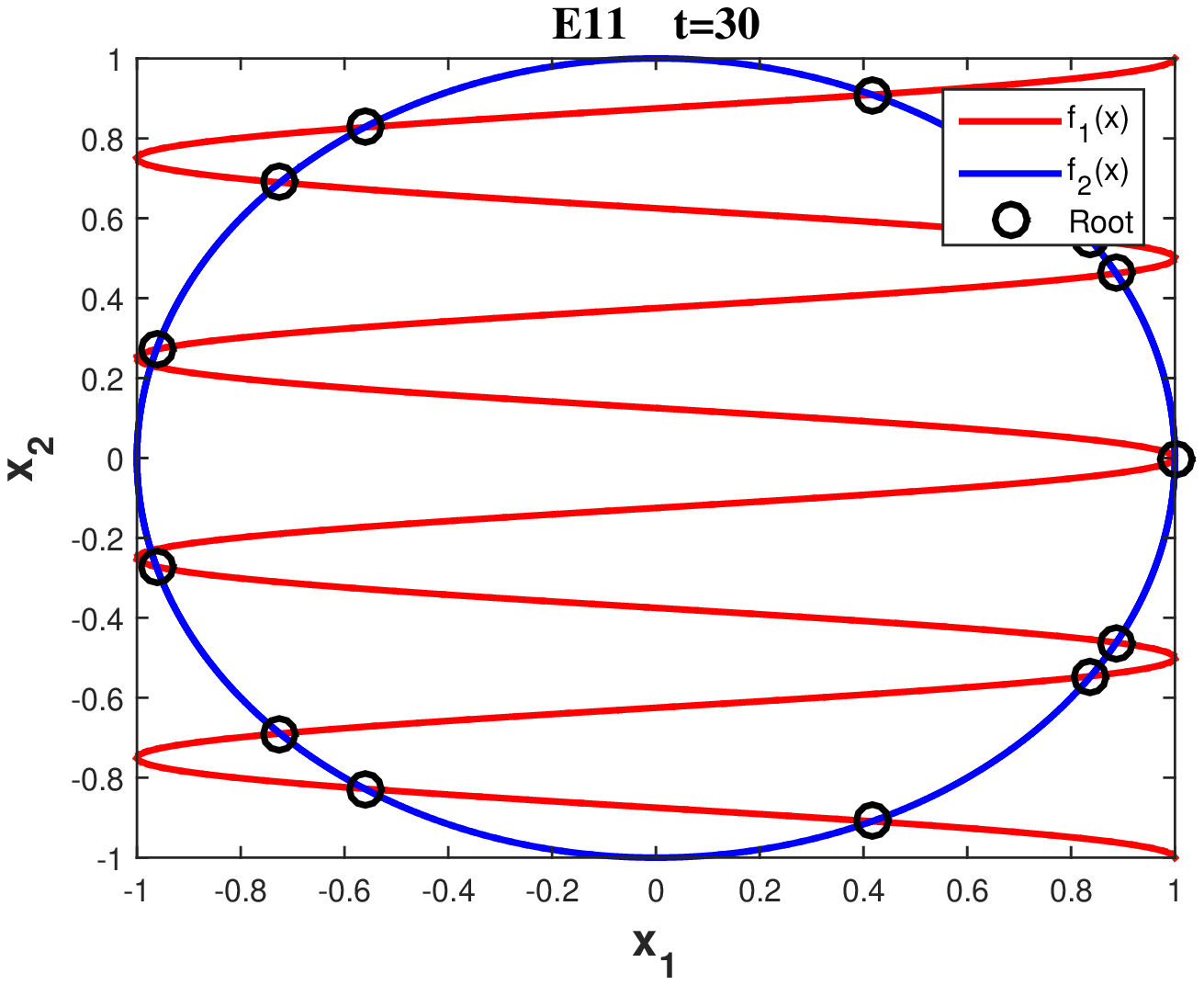,width=1.55in}}
	\end{center}\vspace{-3mm}
	\caption{Evolution of DR-JADE over a typical run on E11} \label{Fig:Evolution of E11}
\end{figure*}

\begin{figure*}[htp]
	\begin{center}
		\subfigure[\small{t=1}]{\psfig{file=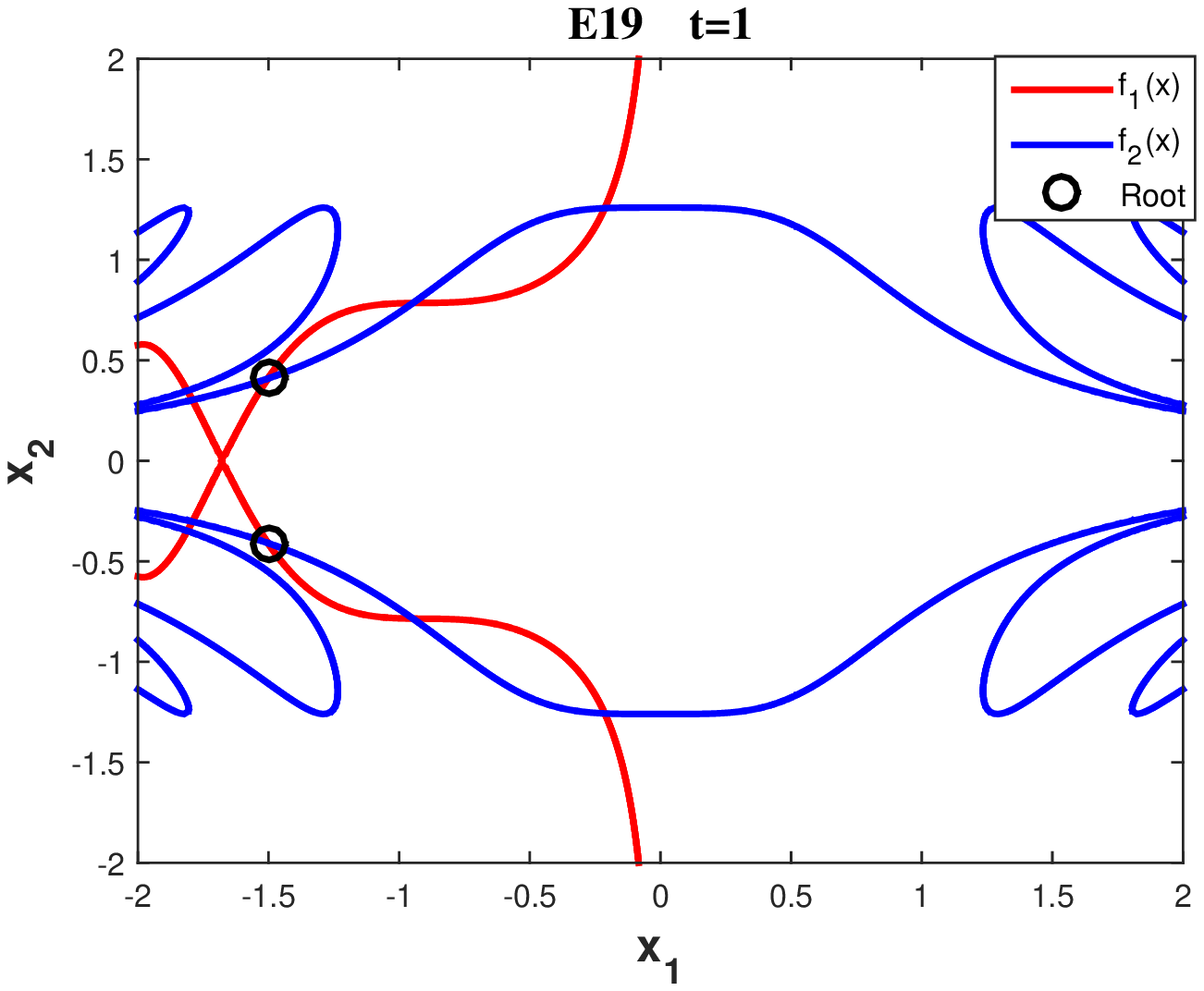,width=1.55in}}
		\subfigure[\small{t=5}]{\psfig{file=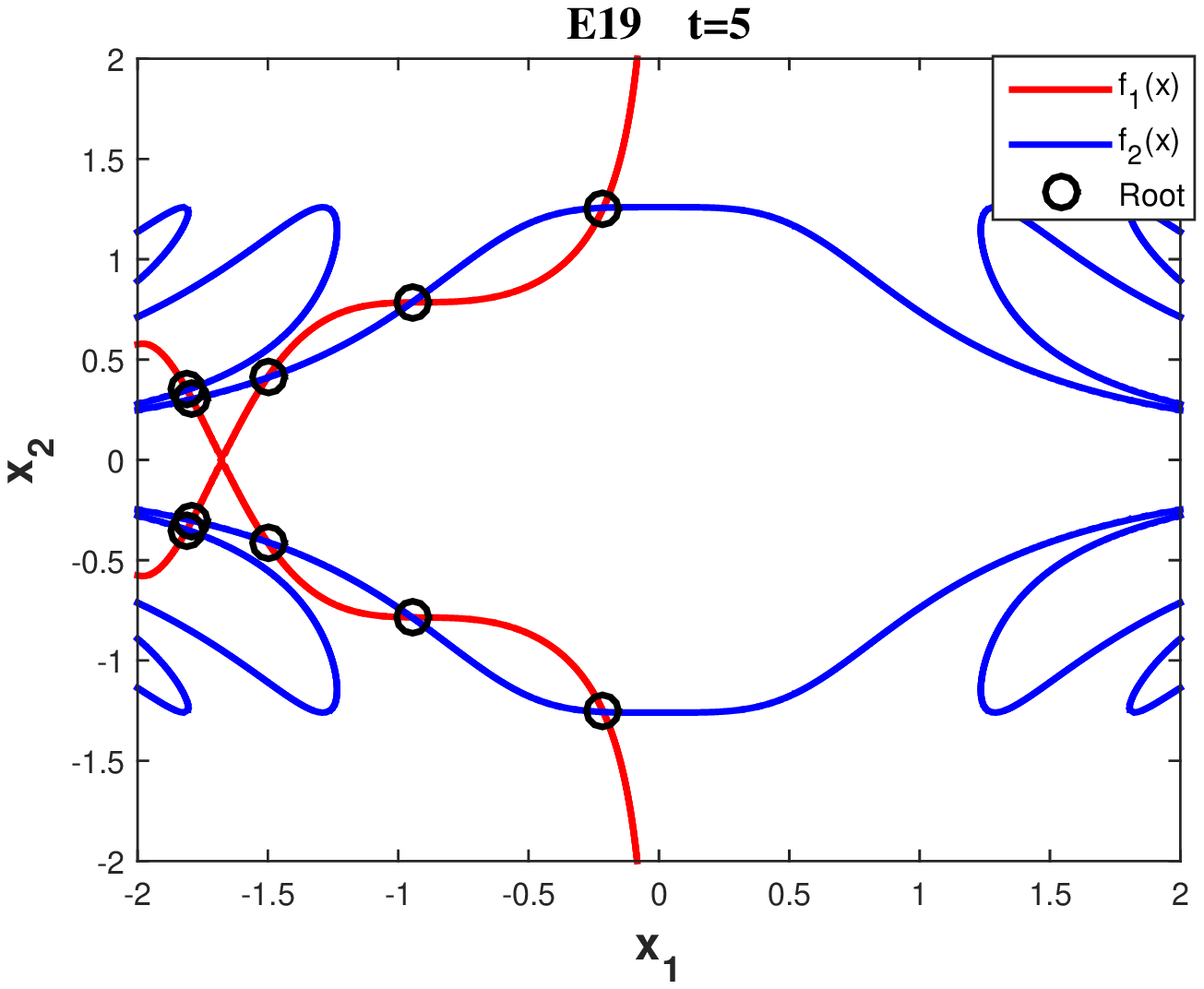,width=1.55in}}
		\subfigure[\small{t=15}]{\psfig{file=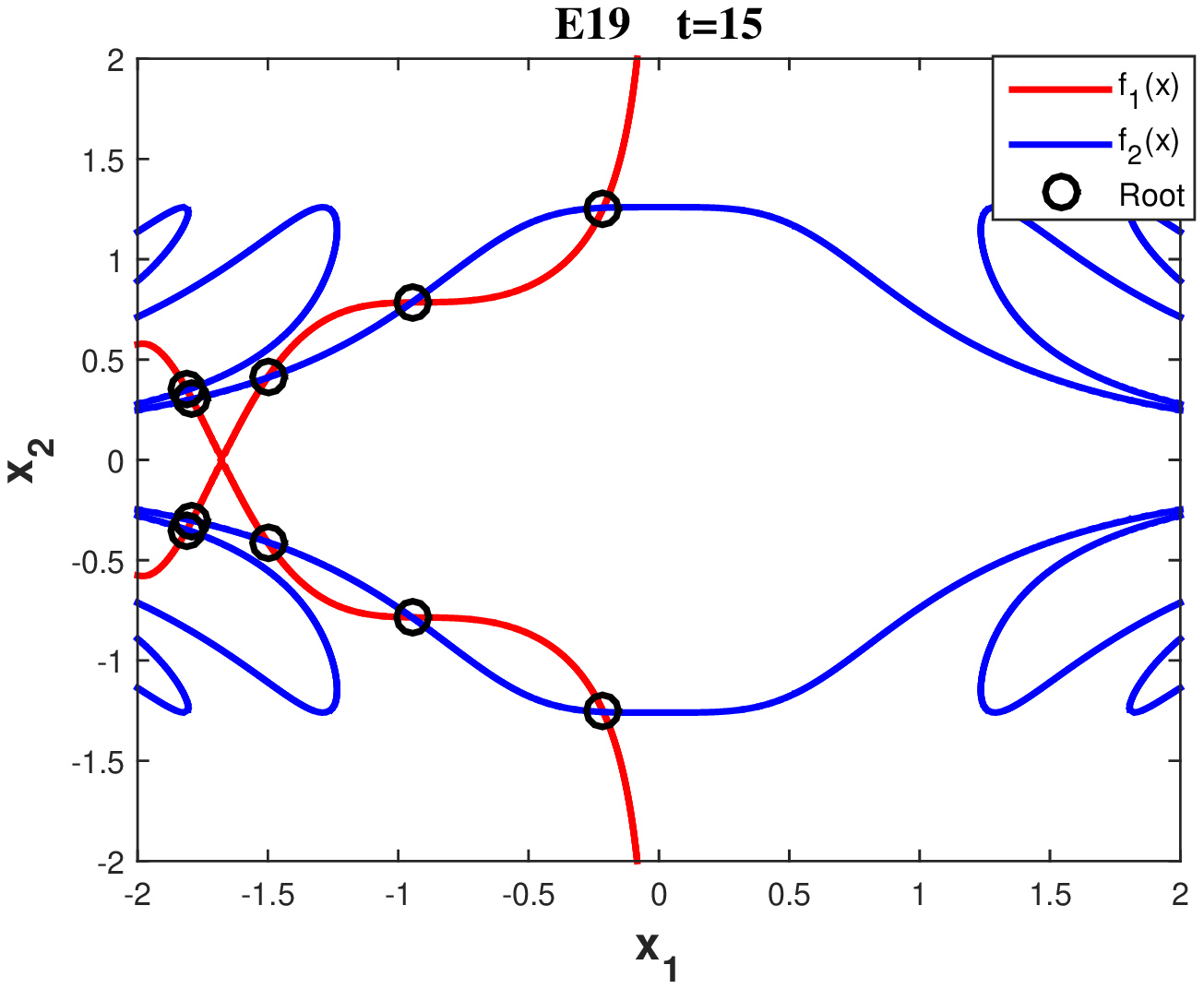,width=1.55in}}
		\subfigure[\small{t=30}]{\psfig{file=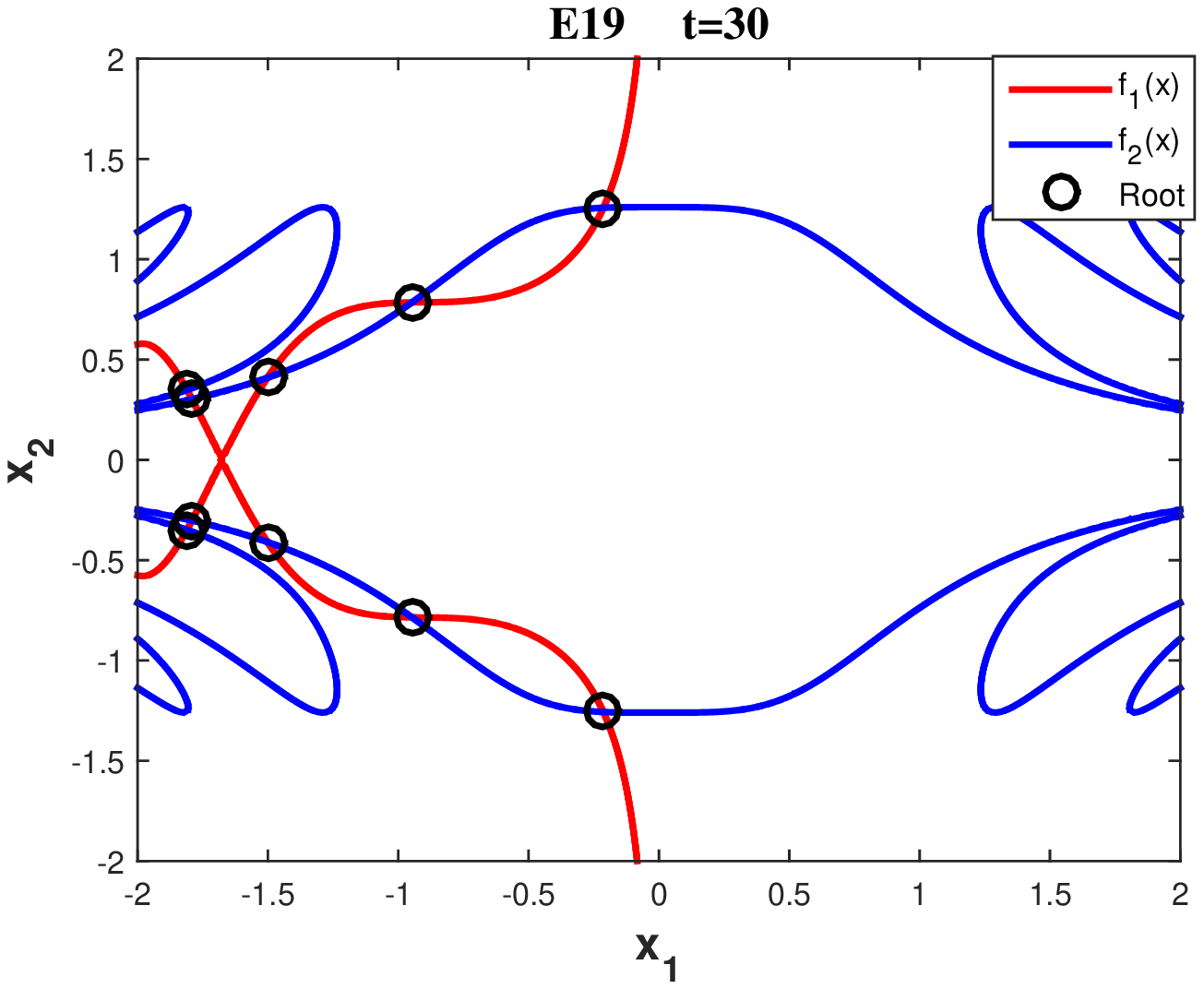,width=1.55in}}
	\end{center}\vspace{-3mm}
	\caption{Evolution of VR-DR-JADE over a typical run on E19} \label{Fig:Evolution of VE19}
\end{figure*}

\begin{figure*}[htp]
	\begin{center}
		\subfigure[\small{t=1}]{\psfig{file=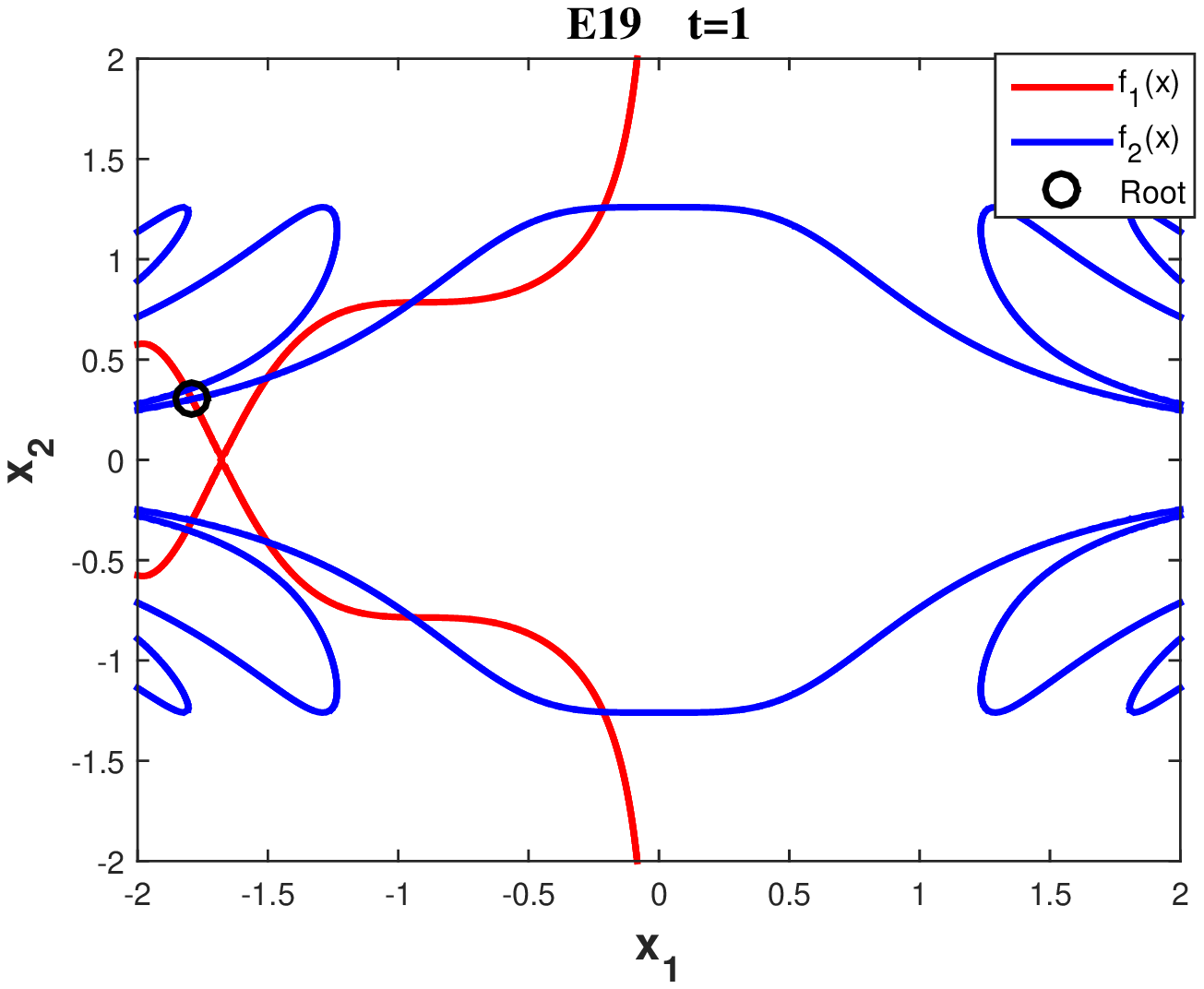,width=1.55in}}
		\subfigure[\small{t=5}]{\psfig{file=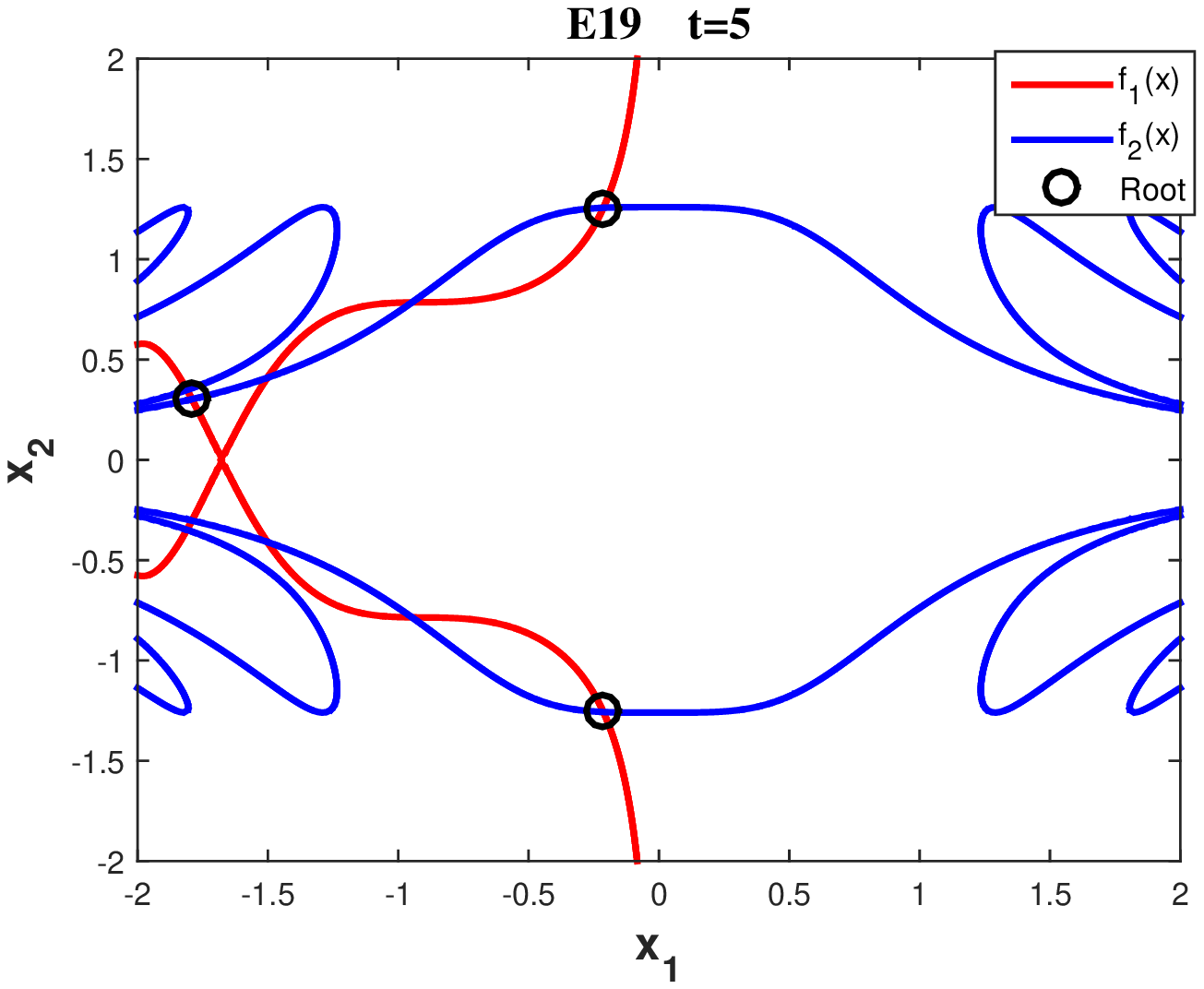,width=1.55in}}
		\subfigure[\small{t=15}]{\psfig{file=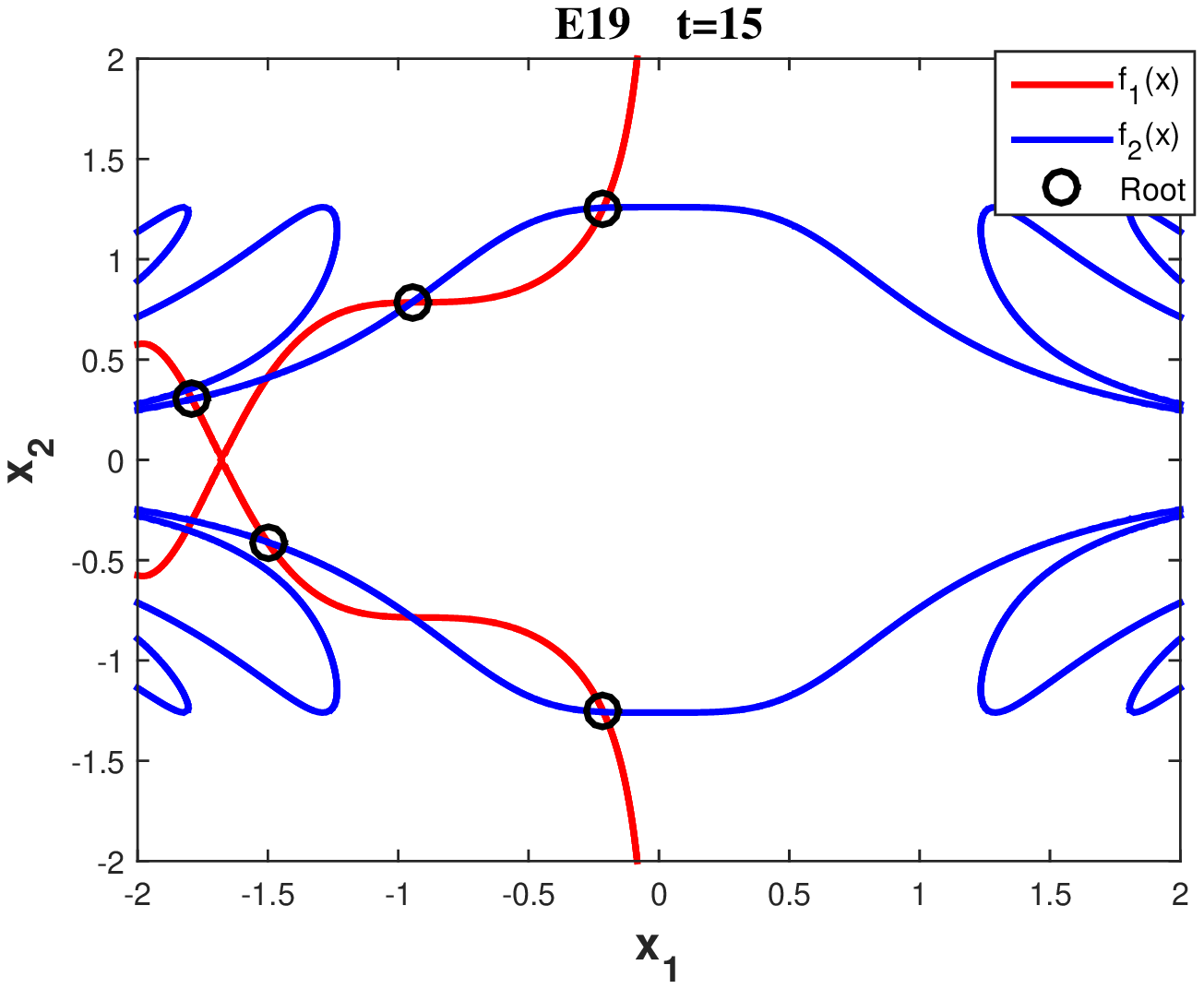,width=1.55in}}
		\subfigure[\small{t=30}]{\psfig{file=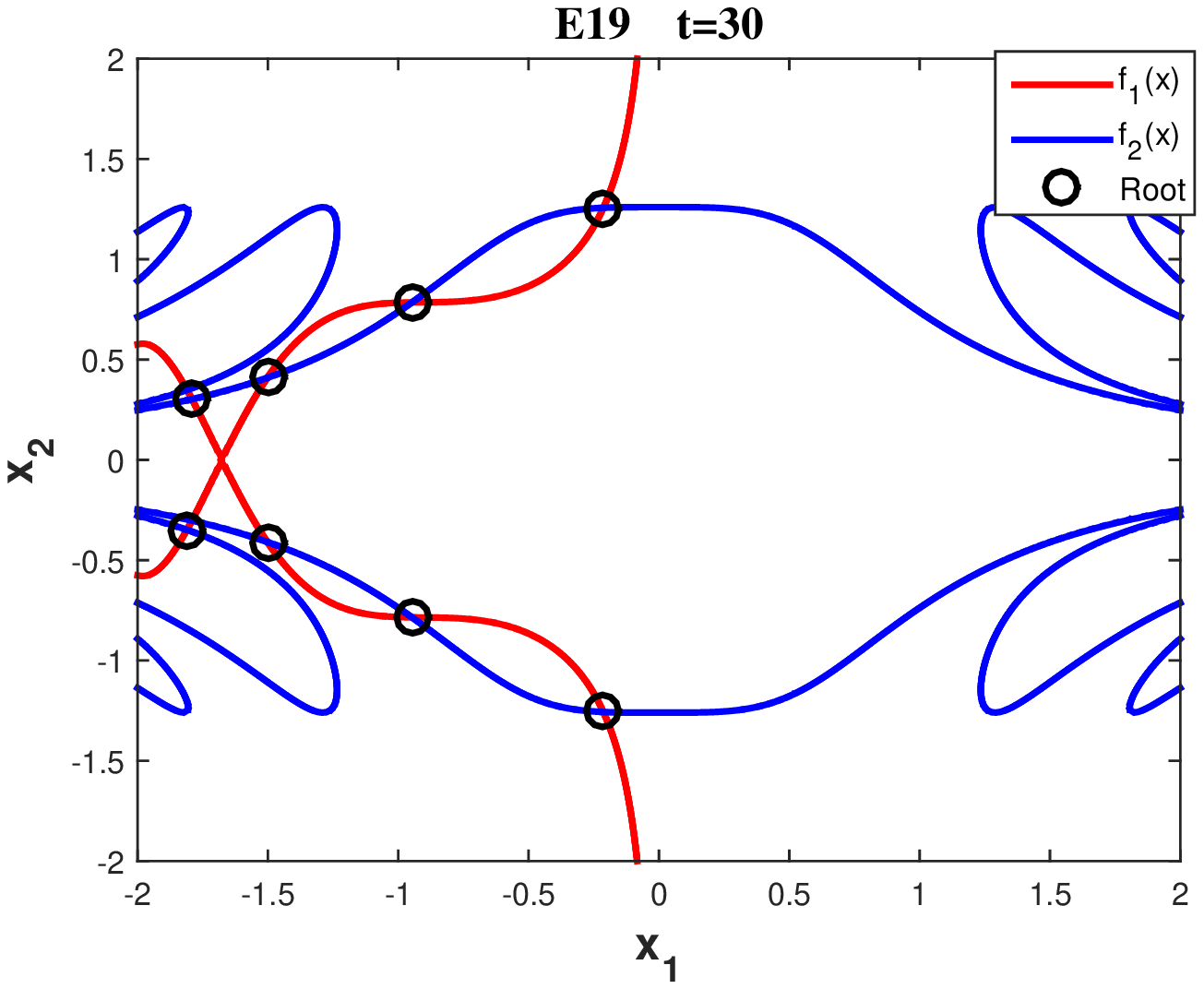,width=1.55in}}
	\end{center}\vspace{-3mm}
	\caption{Evolution of DR-JADE over a typical run on E19} \label{Fig:Evolution of E19}
\end{figure*}

To compare the convergence process of VR-DR-JADE and DR-JADE, we portray the roots located by VR-DR-JADE and DR-JADE when the number of iterations are 1, 5, 15, and 30 over a typical run. Fig.~\ref{Fig:Evolution of VE11}-\ref{Fig:Evolution of E11} and Fig.~\ref{Fig:Evolution of VE19}-\ref{Fig:Evolution of E19} respectively show the results for E11 and E19.

Compared Fig.~\ref{Fig:Evolution of VE11} with Fig.~\ref{Fig:Evolution of E11}, at $t = 1$ VR-DR-JADE and DR-JADE can both find one of the roots for E11. At $t = 5$, VR-DR-JADE can locate five roots while DR-JADE only finds three roots. At $t = 15$, VR-DR-JADE can locate 11 roots, but DR-JADE only locates nine roots. At $t = 30$, VR-DR-JADE has found all 15 roots of E11, while DR-JADE only located 13 roots.

Compared Fig.~\ref{Fig:Evolution of VE19} with Fig.~\ref{Fig:Evolution of E19}, at $t = 1$ owing to the reduced variable with a quadratic term for E19, we can locate two roots for VR-DR-JADE. At $t = 5$, VR-DR-JADE has located all the ten roots while DR-JADE only has found three roots. At $t = 15$ and $t = 30$, DR-JADE has located 5 and 8 roots respectively and there have been still two roots not found at the end of evolution. The above comparison results demonstrate that the application of VRS improves the search efficiency of DR-JADE and makes VR-DR-JADE can get more roots than DR-JADE under the same number of iterations.


In the previous sections, the performance of VR-DR-JADE is verified through the 42 NESs with known roots. For E43-E46, we evaluate the number of obtained roots for DR-JADE and VR-DR-JADE by the best, the worst, the mean, and the standard deviation of the number of obtained roots over 30 runs. What's more, we evaluate the quality of obtained roots for DR-JADE and VR-DR-JADE by the best, the mean, and the standard deviation of objective function values of obtained roots over a typical run. The results are reported in Table~\ref{Table:results of E44-46}\footnote{The date of DR-JADE comes from the literature \cite{liao2018solving}.}.

\aboverulesep=0pt \belowrulesep=0pt
\begin{table*}[htp]
	\footnotesize
	\centering
	\caption{Status of DR-JADE and VR-DR-JADE for the number of obtained roots and the objective function values}
	\label{Table:results of E44-46}
	\begin{center}
		\begin{tabular}{ccccccccc} \toprule
			\multirow{2}{*}{Test problem} & \multirow{2}{*}{Algorithm} & \multicolumn{3}{c}{Objective function values of obtained roots} & \multicolumn{4}{c}{Number of obtained roots}  \\
			\cline{3-9}
			& & Best & Mean & Std & Best & Worst & Mean & Std  \\
			\midrule

			\multirow{2}{*}{E43} & DR-JADE & 1.66E-25 & 3.45E-06 & 3.27E-06 & 12.00 & 7.00 & 8.93 &  1.86  \\
			
			& VR-DR-JADE & {\bf 1.05E-33} & {\bf 1.88E-07} & {\bf 6.37E-07} & {\bf 30.00} & {\bf 30.00} & {\bf 30.00} & {\bf 0.00}  \\ \hline

			\multirow{2}{*}{E44} & DR-JADE & 3.85E-28 & 2.06E-12 & 7.70E-12 & 28.00 & 22.00 & 24.10 & 1.54  \\
			
			& VR-DR-JADE & {\bf 0.00E+00} & {\bf 6.16E-34} & {\bf 3.38E-33} & {\bf 30.00} & {\bf 29.00} & {\bf 29.70} & {\bf 0.47}  \\ \hline

			\multirow{2}{*}{E45} & DR-JADE & 8.45E-17 & 4.07E-13 & 1.36E-12 & 30.00 & 30.00 & 30.00 & 0.00  \\
			
			& VR-DR-JADE & {\bf 9.28E-26} & {\bf 8.95E-16} & {\bf 1.28E-15} & {\bf 30.00} & {\bf 30.00} & {\bf 30.00} & {\bf 0.00}  \\ \hline

			\multirow{2}{*}{E46} & DR-JADE & 2.49E-29 & 4.07E-13 & 1.36E-12 & 4.00 & 3.00 & 3.13 & 0.12  \\
			
			& VR-DR-JADE & {\bf 1.70E-166} & {\bf 1.51E-114} & {\bf 8.13E-114} & {\bf 30.00} & {\bf 28.00} & {\bf 28.97} & 0.72  \\ 
			
			\bottomrule

		\end{tabular}
	\end{center}
\end{table*}

As shown in Table~\ref{Table:results of E44-46}, for each NES in E43-E46, the number of roots obtained by VR-DR-JADE is better than or equivalent to DR-JADE in terms of the best, the worst, the mean, and the standard deviation values (except the standard deviation values of E46). Especially for E43 and E46, the mean values of the number of roots obtained by VR-DR-JADE are 30 and 28.97 over 30 runs respectively, which is much more than those obtained by DR-JADE. For E46, although the standard deviation value of the number of roots obtained by VR-DR-JADE is larger than that obtained by DR-JADE, the best, the worst, and the mean values of the number of roots obtained by VR-DR-JADE have a great improvement. Especially for E44 and E46, the roots found by VR-DR-JADE have significantly better quality than those found by DR-JADE. The above phenomenon reveals that VR-DR-JADE is capable of locating more better roots than DR-JADE when a NES has an infinite number of solutions.

\subsection{Experimental conclusions}\label{s43}

According to the experimental results and discussions above, we can safely draw some conclusions:
\begin{enumerate}
	
	\item VR-MONES can achieve better IGD-indicator and NOF-indicator values than MONES.
	\item For solving NESs E1-E42 with known optimal solutions, VR-DR-JADE almost always can get higher or equal RR-indicator and SR-indicator values than DR-JADE and the other nine methods, and VR-DR-JADE can obtain smaller QR-indicator than DR-JADE.
	\item For solving NESs E43-E46 with unknown optimal solutions, VR-DR-JADE can locate more roots and the obtained roots are closer to the actual roots in a single run compared to DR-JADE.
	
\end{enumerate}

In conclusion, the integration of VRS enables MONES and DR-JADE not only to locate more roots but also significantly improve the quality of the located roots.

\section{Conclusions and Future Work}\label{s5}

This paper proposes to incorporate VRS into EAs to solve NESs. VRS reduces the number of variables and equations of a NES, accordingly shrinks the decision space and reduces the complexity of the NES, and results in improving the optimization efficiency of the original EA for solving the NES.VRS is specifically integrated with two state-of-the-art methods (MONES and DR-JADE), respectively. The experimental results on two test suites with 7 NESs and 46 NESs respectively verify the effectiveness of VRS in solving NESs. According to the framework of the combination of VRS and EAs, VRS theoretically can also be integrated with any EA. The research in this paper reveals that combining with problem domain knowledge could improve the performance of algorithms.

It is noted that there are still shortcomings in this work. On the one hand, VRS cannot be applied to all NESs. On the other hand, for two NESs, EAs may even obtain worse results after integrating with VRS. For the NESs that cannot be explicitly reduced, an approximative variable reduction strategy may be useful to resolve this problem. Moreover, we can develop more efficient transformation techniques and EAs to combine with VRS to solve NESs, e.g., the ensemble algorithms \cite{wu2019ensemble} and objective space partition strategy \cite{chen2019adaptive}. In summary, extending VRS and designing more efficient EAs and transformation techniques to integrate with VRS deserve further investigation in the future.


%
%

\ifCLASSOPTIONcaptionsoff
  \newpage
\fi

\bibliographystyle{IEEEtran}
\bibliography{IEEEabrv,mybib}

\begin{thebibliography}{10}
\providecommand{\url}[1]{#1}
\csname url@samestyle\endcsname
\providecommand{\newblock}{\relax}
\providecommand{\bibinfo}[2]{#2}
\providecommand{\BIBentrySTDinterwordspacing}{\spaceskip=0pt\relax}
\providecommand{\BIBentryALTinterwordstretchfactor}{4}
\providecommand{\BIBentryALTinterwordspacing}{\spaceskip=\fontdimen2\font plus
\BIBentryALTinterwordstretchfactor\fontdimen3\font minus
  \fontdimen4\font\relax}
\providecommand{\BIBforeignlanguage}[2]{{%
\expandafter\ifx\csname l@#1\endcsname\relax
\typeout{** WARNING: IEEEtran.bst: No hyphenation pattern has been}%
\typeout{** loaded for the language `#1'. Using the pattern for}%
\typeout{** the default language instead.}%
\else
\language=\csname l@#1\endcsname
\fi
#2}}
\providecommand{\BIBdecl}{\relax}
\BIBdecl

\bibitem{mehta2015collection}
D.~Mehta and C.~Grosan, ``A collection of challenging optimization problems in
  science, engineering and economics,'' in \emph{2015 IEEE Congress on
  Evolutionary Computation (CEC)}.\hskip 1em plus 0.5em minus 0.4em\relax IEEE,
  2015, pp. 2697--2704.

\bibitem{holstad1999numerical}
A.~Holstad, ``Numerical solution of nonlinear equations in chemical speciation
  calculations,'' \emph{Computational geosciences}, vol.~3, no. 3-4, pp.
  229--257, 1999.

\bibitem{collins2002forward}
C.~L. Collins, ``Forward kinematics of planar parallel manipulators in the
  clifford algebra of p2,'' \emph{Mechanism and Machine Theory}, vol.~37,
  no.~8, pp. 799--813, 2002.

\bibitem{facchinei2007generalized}
F.~Facchinei and C.~Kanzow, ``Generalized nash equilibrium problems,''
  \emph{4or}, vol.~5, no.~3, pp. 173--210, 2007.

\bibitem{chaudhary2017modified}
N.~I. Chaudhary, M.~S. Aslam, and M.~A.~Z. Raja, ``Modified volterra lms
  algorithm to fractional order for identification of hammerstein non-linear
  system,'' \emph{IET Signal Processing}, vol.~11, no.~8, pp. 975--985, 2017.

\bibitem{yuan2008new}
G.~Yuan and X.~Lu, ``A new backtracking inexact bfgs method for symmetric
  nonlinear equations,'' \emph{Computers \& Mathematics with Applications},
  vol.~55, no.~1, pp. 116--129, 2008.

\bibitem{karr1998solutions}
C.~L. Karr, B.~Weck, and L.~M. Freeman, ``Solutions to systems of nonlinear
  equations via a genetic algorithm,'' \emph{Engineering Applications of
  Artificial Intelligence}, vol.~11, no.~3, pp. 369--375, 1998.

\bibitem{grosan2008new}
C.~Grosan and A.~Abraham, ``A new approach for solving nonlinear equations
  systems,'' \emph{IEEE Transactions on Systems, Man, and Cybernetics-Part A:
  Systems and Humans}, vol.~38, no.~3, pp. 698--714, 2008.

\bibitem{bates2013numerically}
D.~J. Bates, A.~J. Sommese, J.~D. Hauenstein, and C.~W. Wampler,
  \emph{Numerically solving polynomial systems with Bertini}.\hskip 1em plus
  0.5em minus 0.4em\relax SIAM, 2013.

\bibitem{denis1993least}
J.~Denis and H.~Wolkowicz, ``Least change secant methods, sizing, and
  shifting,'' \emph{SIAM Journal of Numerical Analisys}, vol.~30, pp.
  1291--1314, 1993.

\bibitem{ortega1970iterative}
J.~M. Ortega and W.~C. Rheinboldt, \emph{Iterative solution of nonlinear
  equations in several variables}.\hskip 1em plus 0.5em minus 0.4em\relax Siam,
  1970, vol.~30.

\bibitem{wang2001intelligent}
L.~Wang, ``Intelligent optimization algorithms with applications,''
  \emph{Tsinghua University \& Springer Press, Beijing}, 2001.

\bibitem{geng2009research}
H.-T. Geng, Y.-J. Sun, Q.-X. Song, and T.-T. Wu, ``Research of ranking method
  in evolution strategy for solving nonlinear system of equations,'' in
  \emph{2009 First International Conference on Information Science and
  Engineering}.\hskip 1em plus 0.5em minus 0.4em\relax IEEE, 2009, pp.
  348--351.

\bibitem{ouyang2009hybrid}
A.~Ouyang, Y.~Zhou, and Q.~Luo, ``Hybrid particle swarm optimization algorithm
  for solving systems of nonlinear equations,'' in \emph{2009 IEEE
  International Conference on Granular Computing}.\hskip 1em plus 0.5em minus
  0.4em\relax IEEE, 2009, pp. 460--465.

\bibitem{turgut2014chaotic}
O.~E. Turgut, M.~S. Turgut, and M.~T. Coban, ``Chaotic quantum behaved particle
  swarm optimization algorithm for solving nonlinear system of equations,''
  \emph{Computers \& Mathematics with Applications}, vol.~68, no.~4, pp.
  508--530, 2014.

\bibitem{gong2018finding}
W.~Gong, Y.~Wang, Z.~Cai, and L.~Wang, ``Finding multiple roots of nonlinear
  equation systems via a repulsion-based adaptive differential evolution,''
  \emph{IEEE Transactions on Systems, Man, and Cybernetics: Systems}, vol.~50,
  no.~4, pp. 1499--1513, 2020.

\bibitem{ren2013solving}
H.~Ren, L.~Wu, W.~Bi, and I.~K. Argyros, ``Solving nonlinear equations system
  via an efficient genetic algorithm with symmetric and harmonious
  individuals,'' \emph{Applied Mathematics and Computation}, vol. 219, no.~23,
  pp. 10\,967--10\,973, 2013.

\bibitem{joshi2014solving}
G.~Joshi and M.~B. Krishna, ``Solving system of non-linear equations using
  genetic algorithm,'' in \emph{2014 International Conference on Advances in
  Computing, Communications and Informatics (ICACCI)}.\hskip 1em plus 0.5em
  minus 0.4em\relax IEEE, 2014, pp. 1302--1308.

\bibitem{beyer2002evolution}
H.-G. Beyer and H.-P. Schwefel, ``Evolution strategies--a comprehensive
  introduction,'' \emph{Natural computing}, vol.~1, no.~1, pp. 3--52, 2002.

\bibitem{wu2015variable}
G.~Wu, W.~Pedrycz, P.~N. Suganthan, and R.~Mallipeddi, ``A variable reduction
  strategy for evolutionary algorithms handling equality constraints,''
  \emph{Applied Soft Computing}, vol.~37, pp. 774--786, 2015.

\bibitem{wu2017using}
G.~Wu, W.~Pedrycz, P.~N. Suganthan, and H.~Li, ``Using variable reduction
  strategy to accelerate evolutionary optimization,'' \emph{Applied Soft
  Computing}, vol.~61, pp. 283--293, 2017.

\bibitem{liao2018solving}
Z.~Liao, W.~Gong, X.~Yan, L.~Wang, and C.~Hu, ``Solving nonlinear equations
  system with dynamic repulsion-based evolutionary algorithms,'' \emph{IEEE
  Transactions on Systems, Man, and Cybernetics: Systems}, vol.~50, no.~4, pp.
  1590--1601, 2020.

\bibitem{song2014locating}
W.~Song, Y.~Wang, H.-X. Li, and Z.~Cai, ``Locating multiple optimal solutions
  of nonlinear equation systems based on multiobjective optimization,''
  \emph{IEEE Transactions on Evolutionary Computation}, vol.~19, no.~3, pp.
  414--431, 2014.

\bibitem{qin2015nonlinear}
S.~Qin, S.~Zeng, W.~Dong, and X.~Li, ``Nonlinear equation systems solved by
  many-objective hype,'' in \emph{2015 IEEE Congress on Evolutionary
  Computation (CEC)}.\hskip 1em plus 0.5em minus 0.4em\relax IEEE, 2015, pp.
  2691--2696.

\bibitem{hirsch2009solving}
M.~J. Hirsch, P.~M. Pardalos, and M.~G. Resende, ``Solving systems of nonlinear
  equations with continuous grasp,'' \emph{Nonlinear Analysis: Real World
  Applications}, vol.~10, no.~4, pp. 2000--2006, 2009.

\bibitem{wang2014mommop}
Y.~Wang, H.-X. Li, G.~G. Yen, and W.~Song, ``Mommop: Multiobjective
  optimization for locating multiple optimal solutions of multimodal
  optimization problems,'' \emph{IEEE transactions on cybernetics}, vol.~45,
  no.~4, pp. 830--843, 2014.

\bibitem{mousa2008GENLS}
A.~Mousa and I.~El-Desoky, ``Genls: Co-evolutionary algorithm for nonlinear
  system of equations,'' \emph{Applied Mathematics and Computation}, vol. 197,
  no.~2, pp. 633--642, 2008.

\bibitem{kuri2003solution}
A.~F. Kuri-Morales, R.~H. No, and D.~M{\'e}xico, ``Solution of simultaneous
  non-linear equations using genetic algorithms,'' \emph{WSEAS Transactions on
  Systems}, vol.~2, no.~1, pp. 44--51, 2003.

\bibitem{deb2002fast}
K.~Deb, A.~Pratap, S.~Agarwal, and T.~Meyarivan, ``A fast and elitist
  multiobjective genetic algorithm: Nsga-ii,'' \emph{IEEE transactions on
  evolutionary computation}, vol.~6, no.~2, pp. 182--197, 2002.

\bibitem{zhang2009JADE}
J.~Zhang and A.~C. Sanderson, ``Jade: adaptive differential evolution with
  optional external archive,'' \emph{IEEE Transactions on evolutionary
  computation}, vol.~13, no.~5, pp. 945--958, 2009.

\bibitem{wang2012regularity}
Y.~Wang, J.~Xiang, and Z.~Cai, ``A regularity model-based multiobjective
  estimation of distribution algorithm with reducing redundant cluster
  operator,'' \emph{Applied Soft Computing}, vol.~12, no.~11, pp. 3526--3538,
  2012.

\bibitem{triguero2017keel}
I.~Triguero, S.~Gonz{\'a}lez, J.~M. Moyano \emph{et~al.}, ``Keel 3.0: an open
  source software for multi-stage analysis in data mining,''
  \emph{International Journal of Computational Intelligence Systems}, vol.~10,
  pp. 1238--1249, 2017.

\bibitem{thomsen2004multimodal}
R.~Thomsen, ``Multimodal optimization using crowding-based differential
  evolution,'' in \emph{Proceedings of the 2004 Congress on Evolutionary
  Computation (IEEE Cat. No. 04TH8753)}, vol.~2.\hskip 1em plus 0.5em minus
  0.4em\relax IEEE, 2004, pp. 1382--1389.

\bibitem{li2013benchmark}
X.~Li, A.~Engelbrecht, and M.~G. Epitropakis, ``Benchmark functions for
  cec’2013 special session and competition on niching methods for multimodal
  function optimization,'' \emph{RMIT University, Evolutionary Computation and
  Machine Learning Group, Australia, Tech. Rep}, 2013.

\bibitem{gong2017weighted}
W.~Gong, Y.~Wang, Z.~Cai, and S.~Yang, ``A weighted biobjective transformation
  technique for locating multiple optimal solutions of nonlinear equation
  systems,'' \emph{IEEE Transactions on Evolutionary Computation}, vol.~21,
  no.~5, pp. 697--713, 2017.

\bibitem{ramadas2014multiple}
G.~C. Ramadas, E.~M. Fernandes, and A.~M.~A. Rocha, ``Multiple roots of systems
  of equations by repulsion merit functions,'' in \emph{International
  Conference on Computational Science and Its Applications}.\hskip 1em plus
  0.5em minus 0.4em\relax Springer, 2014, pp. 126--139.

\bibitem{qu2012differential}
B.-Y. Qu, P.~N. Suganthan, and J.-J. Liang, ``Differential evolution with
  neighborhood mutation for multimodal optimization,'' \emph{IEEE transactions
  on evolutionary computation}, vol.~16, no.~5, pp. 601--614, 2012.

\bibitem{raja2016memetic}
M.~A.~Z. Raja, A.~K. Kiani, A.~Shehzad, and A.~Zameer, ``Memetic computing
  through bio-inspired heuristics integration with sequential quadratic
  programming for nonlinear systems arising in different physical models,''
  \emph{SpringerPlus}, vol.~5, no.~1, p. 2063, 2016.

\bibitem{raja2018nature}
M.~A.~Z. Raja, A.~Zameer, A.~K. Kiani, A.~Shehzad, and M.~A.~R. Khan,
  ``Nature-inspired computational intelligence integration with nelder--mead
  method to solve nonlinear benchmark models,'' \emph{Neural Computing and
  Applications}, vol.~29, no.~4, pp. 1169--1193, 2018.

\bibitem{zhang2019applying}
X.~Zhang, Q.~Wan, and Y.~Fan, ``Applying modified cuckoo search algorithm for
  solving systems of nonlinear equations,'' \emph{Neural Computing and
  Applications}, vol.~31, no.~2, pp. 553--576, 2019.

\bibitem{wu2019ensemble}
G.~Wu, R.~Mallipeddi, and P.~N. Suganthan, ``Ensemble strategies for
  population-based optimization algorithms--a survey,'' \emph{Swarm and
  evolutionary computation}, vol.~44, pp. 695--711, 2019.

\bibitem{chen2019adaptive}
H.~Chen, G.~Wu, W.~Pedrycz, P.~N. Suganthan, L.~Xing, and X.~Zhu, ``An adaptive
  resource allocation strategy for objective space partition-based
  multiobjective optimization,'' \emph{IEEE Transactions on Systems, Man, and
  Cybernetics: Systems}, pp. 1--16, 2019.

\end{thebibliography}

%

\end{document}